\documentclass[letterpaper, 10 pt, conference]{ieeeconf}
\usepackage{subcaption}
\IEEEoverridecommandlockouts

\usepackage{cite}
\usepackage{amsmath,amssymb,amsfonts}
\usepackage{algorithmic}
\usepackage{afterpage}
\usepackage{algorithm}
\usepackage{subcaption}
\usepackage{mdframed}
\usepackage{graphicx}
\usepackage{textcomp}
\usepackage{xcolor}
\usepackage{tabularx}
\usepackage{array}
\usepackage{subfig}
\usepackage{textcomp}
\usepackage{stfloats}
\usepackage{verbatim}
\usepackage{booktabs}
\usepackage{multirow}
\usepackage{url}
\usepackage{multicol}
\usepackage{bm}

\def\BibTeX{{\rm B\kern-.05em{\sc i\kern-.025em b}\kern-.08em
    T\kern-.1667em\lower.7ex\hbox{E}\kern-.125emX}}
\begin{document}

\title{\LARGE \bf
 Distributional Soft Actor-Critic with Diffusion Policy }

\author{
    Tong Liu$^{1*}$, 
    Yinuo Wang$^{1*}$, 
    Xujie Song$^{1*}$, 
    Wenjun Zou$^{1}$, 
    Liangfa Chen$^{2}$, 
    Likun Wang$^{1}$, 
    Bin Shuai$^{1}$, 
    \\
    Jingliang Duan$^{2}$, 
    Shengbo Eben Li$^{1}$
    \thanks{
        $^{1}$Tong Liu, Yinuo Wang, Xujie Song, Wenjun Zou, Likun Wang, Bin Shuai and Shengbo Eben Li are with the School of Vehicle and Mobility, Tsinghua University, Beijing 100084, China.
        (Emails: \texttt{\{liu-t22, wyn23, songxj21, zouwj20, wlk23\}@mails.tsinghua.edu.cn}, \texttt{shuaib@mail.tsinghua.edu.cn}, \texttt{lishbo@tsinghua.edu.cn})
    }
    \thanks{
        $^{2}$Liangfa Chen, Jingliang Duan are with the School of Mechanical Engineering, University of Science and Technology Beijing, Beijing 100083, China.
        (Emails: \texttt{chenliangfa@xs.ustb.edu.cn}, \texttt{duanjl@ustb.edu.cn})
    }
        \thanks{ 
        This work was supported by National Key Research and Development Program of China under grant number 2022YFB2502901.\textit{Corresponding author: Shengbo Eben Li (lishbo@tsinghua.edu.cn)}
    }
}

\maketitle
\begin{abstract}
Reinforcement learning has been proven to be highly effective in handling complex control tasks. Traditional methods typically use unimodal distributions, such as Gaussian distributions, to model the output of value distributions. However, unimodal distribution often and easily causes bias in value function estimation, leading to poor algorithm performance. This paper proposes a distributional reinforcement learning algorithm called DSAC-D (Distributed Soft Actor Critic with Diffusion Policy) to address the challenges of estimating bias in value functions and obtaining multimodal policy representations. A multimodal distributional policy iteration framework that can converge to the optimal policy was established by introducing policy entropy and value distribution function. A diffusion value network that can accurately characterize the distribution of multi peaks was constructed by generating a set of reward samples through reverse sampling using a diffusion model. Based on this, a distributional reinforcement learning algorithm with dual diffusion of the value network and the policy network was derived. MuJoCo testing tasks demonstrate that the proposed algorithm not only learns multimodal policy, but also achieves state-of-the-art (SOTA) performance in all 9 control tasks, with significant suppression of estimation bias and total average return improvement of over 10\% compared to existing mainstream algorithms. The results of real vehicle testing show that DSAC-D can accurately characterize the multimodal distribution of different driving styles, and the diffusion policy network can characterize multimodal trajectories.
\end{abstract}
\section{INTRODUCTION}
Autonomous robotics technology is a significant intelligent automation advancement with great potential for enhancing safety, efficiency, and adaptability in various fields \cite{kakolu2023autonomous, xianjia2021applications}. Yet, its practical deployment is challenging, especially in dynamic and unstructured environments where robots must make critical decisions like obstacle avoidance.

Reinforcement learning (RL) is crucial for decision-making in autonomous robotics \cite{hua2023recent}. Distributional Reinforcement Learning, a key RL advancement, models the full probability distribution of cumulative returns, capturing uncertainties and enhancing policy robustness \cite{bellemare2017distributional}. C51 started discrete distribution modeling with 51 fixed atomic values and KL divergence. QR-DQN increased flexibility via quantile regression \cite{dabney2018implicit}. IQN achieved continuous modeling through neural-network-based adaptive sampling, and FQF refined representation by adjusting quantile positions \cite{yang2019fully}. Distributional Reinforcement Learning has great potential for improving stability and decision-making.

In real-world environments, value distributions are often multimodal. Assuming a unimodal distribution causes significant loss of distributional information, leading to value estimation bias, suboptimal policy learning, and poor adaptability to complex scenarios. Thus, the above algorithms fail to capture the complexities in autonomous driving scenarios. 

To address these challenges, this paper proposes a distributed reinforcement learning algorithm called DSAC-D (Distributed Soft Actor Critic with Diffusion Policy) to reduce value function estimation bias and obtain multimodal policy representations. The contributions of this paper are as follows:

\begin{itemize}
    \item We propose a diffusion value network (DVN) that can accurately characterize the value distribution of multiple peaks by generating a set of reward samples through reverse sampling using a diffusion model. This breaks through the limitations of traditional unimodal value distribution and greatly suppresses value estimation bias. 
    \item We propose a distributional reinforcement learning algorithm DSAC-D. A multimodal distributional policy iteration framework that can converge to the optimal policy was established by introducing policy entropy and value distribution function. Based on this, a distributional reinforcement learning algorithm with dual diffusion of the value network and the policy network was derived. 
    \item We integrate the diffusion policy as an approximate function module, called the DiffusionNet. Experiments conducted on MuJoCo benchmarks show that DSAC-D not only facilitates the multimodal policy representation capability but also achieves state-of-the-art performance. In the stochastic vehicle meeting environment, DSAC-D can learn different multi-modal Q-value distributions and output multimodal trajectories. 
\end{itemize}

\section{Preliminaries}
\label{sec:pre}
\subsection{Online Reinforcement Learning}
\label{sec:pre_onlineRL}
Reinforcement Learning (RL) is usually modeled as a Markov Decision Process (MDP), offering a mathematical way to make decisions in stochastic environments. The aim of RL is to find an optimal policy \(\pi\) that maximizes the expected cumulative discounted return \(J(\pi)\). The state-action value function \(Q^\pi(s, a)\) estimates the expected return of taking action \(a\) in state \(s\) under policy \(\pi\):
\begin{equation}
    Q^\pi(s, a) = \mathbb{E} \left[ \sum_{t=0}^{\infty} \gamma^t R_t \mid s, a, \pi \right].
\end{equation}

In online RL, an agent interacts with the environment iteratively, updating its policy based on observed state transitions and rewards. A widely used approach is the actor-critic framework, which alternates between policy evaluation and policy improvement. During policy evaluation, the Q-value function is refined according to the Bellman equation:  
\begin{equation}
Q^\pi(s, a) \leftarrow R(s, a) + \gamma \mathbb{E}_{s' \sim P, a' \sim \pi}[Q^\pi(s', a')].
\end{equation} 

The policy improvement step aims to maximize the expected Q-value, often using a greedy update strategy:  
\begin{equation}
\pi_{\text{new}}(\cdot | s) = \arg \max_{\pi \in \Pi} \mathbb{E}_{a \sim \pi} [Q^{\pi_{\text{old}}} (s, a)].
\end{equation} 

Through iterative refinement, the agent gradually improves its decision-making process, ultimately converging toward an optimal policy.

\subsection{Diffusion Models}
Diffusion models are a type of generative model that perform well in creating high-dimensional data, like images and audio \cite{ho2020denoising}. These models are built upon a two-step stochastic process: forward diffusion process and reverse generative process. This formulation enables diffusion models to capture complex multimodal distributions, making them effective for high-quality generative modeling.

The forward process is formulated as a Markov chain that incrementally adds Gaussian noise to the data sample \( x_0 \) over \( T \) timesteps. Given a variance schedule \( \{\beta_t\}_{t=1}^{T} \), where \( \beta_t \in (0,1) \), the transition probabilities are defined as:
\begin{equation}
    q(\boldsymbol{x}_{1:T}|\boldsymbol{x}_0) =  \prod_{t=1}^{T} q(\boldsymbol{x}_t|\boldsymbol{x}_{t-1}),
\end{equation}  
\begin{equation}
    q(\boldsymbol{x}_t|\boldsymbol{x}_{t-1}) = \mathcal{N} (\boldsymbol{x}_t; \sqrt{1-\beta_t} \boldsymbol{x}_{t-1}, \beta_t \mathbf{I}).
\end{equation}


To generate new samples, diffusion models define a learnable reverse Markov chain that attempts to recover the original data from noise. The reverse process is parameterized as:
\begin{equation}
    p_{\theta}(\boldsymbol{x}_{0:T}) = p(\boldsymbol{x}_T) \prod_{t=1}^{T} p_{\theta}(\boldsymbol{x}_{t-1}|\boldsymbol{x}_t),
\end{equation}
where \( p(\boldsymbol{x}_T) = \mathcal{N}(\boldsymbol{x}_T; 0, \mathbf{I}) \) is the prior distribution. Each transition in the reverse process is modeled as a Gaussian:
\begin{equation}
    p_{\theta}(\boldsymbol{x}_{t-1}|\boldsymbol{x}_t) = \mathcal{N}(\boldsymbol{x}_{t-1}; \mu_{\theta}(\boldsymbol{x}_t, t), \Sigma_{\theta}(\boldsymbol{x}_t, t)).
\end{equation}
Here, \( \mu_{\theta}(\bm{x}_t, t) \) and \( \Sigma_{\theta}(\bm{x}_t, t) \) are learned functions, typically parameterized by a deep neural network, which predict the mean and variance of the denoised estimate at each step.

\section{Method}
\subsection{Multimodal Distributional Policy Iteration Framework}
We define the stochastic cumulative return \(Z^{\pi}(s, a)\) generated by the policy \(\pi\) starting from the state-action pair \((s, a)\) as:
\begin{equation}
Z^{\pi}\left(s_{i}, a_{i}\right) = r_{i}+\sum_{j = 1}^{\infty} \gamma^{j}\left[r_{i + j}-\alpha\log\pi\left(a_{i + j}\mid s_{i + j}\right)\right],
\end{equation}
Here, \(s\) represents the state of the environment, \(a\) (where \(a_{i>i} \sim \pi\)) represents the action taken in that state, and \(\pi\) is the policy function.

We define its probability density function as \(\mathcal{Z}^{\pi}(\cdot\mid s, a)\), i.e., \(Z^{\pi}(s, a) \sim \mathcal{Z}^{\pi}(\cdot\mid s, a)\), which represents the probability distribution of the cumulative return \(Z\) for the state-action pair \((s, a)\) under the policy \(\pi\).

The multimodal distributional function satisfies the following self-consistency condition:
\begin{equation}
Z^{\pi}(s, a)\stackrel {D}{=} r+\gamma\left[Z\left(s', a'\right)-\alpha\log\pi\left(a'\mid s'\right)\right],
\end{equation}
where \(X \stackrel{D}{=} Y\) means the probability density functions of random variables \(X\) and \(Y\) are equal. We call this process the multimodal distributional policy evaluation step.

The objective of multimodal distributional policy optimization can be rewritten as:
\begin{equation}
\begin{aligned}
\max_{\pi} J(\pi)&=\mathbb{E}_{(s, a) \sim \rho_{\pi}}\left[\mathbb{E}_{Z^{\pi}(s, a) \sim \mathcal{Z}^{\pi}(\cdot\mid s, a)}\left[Z^{\pi}(s, a)\right]\right.\\
&\quad\left.-\alpha\log\pi(a\mid s)\right].
\end{aligned}
\end{equation}

Given the current policy \(\pi_{\text{old}}\), the goal of policy improvement is to find a new policy \(\pi_{\text{new}}\) such that \(J(\pi_{\text{new}}) \geq J(\pi_{\text{old}})\). The corresponding multimodal distributional policy improvement step is:
\begin{equation}
\begin{aligned}
\pi_{\text{new}}&=\arg\max_{\pi}\mathbb{E}\left[\mathbb{E}_{Z^{\pi_{\text{old}}}(s, a) \sim \mathcal{Z}^{\pi_{\text{old}}}(\cdot\mid s, a)}\left[Z^{\pi_{\text{old}}}(s, a)\right]\right.\\
&\quad\left.-\alpha\log\pi_{\text{old}}(a\mid s)\right].
\end{aligned}
\end{equation}

We call this process the multimodal distributional policy improvement step.

\begin{figure}
\centering
\includegraphics[width=0.45\textwidth]{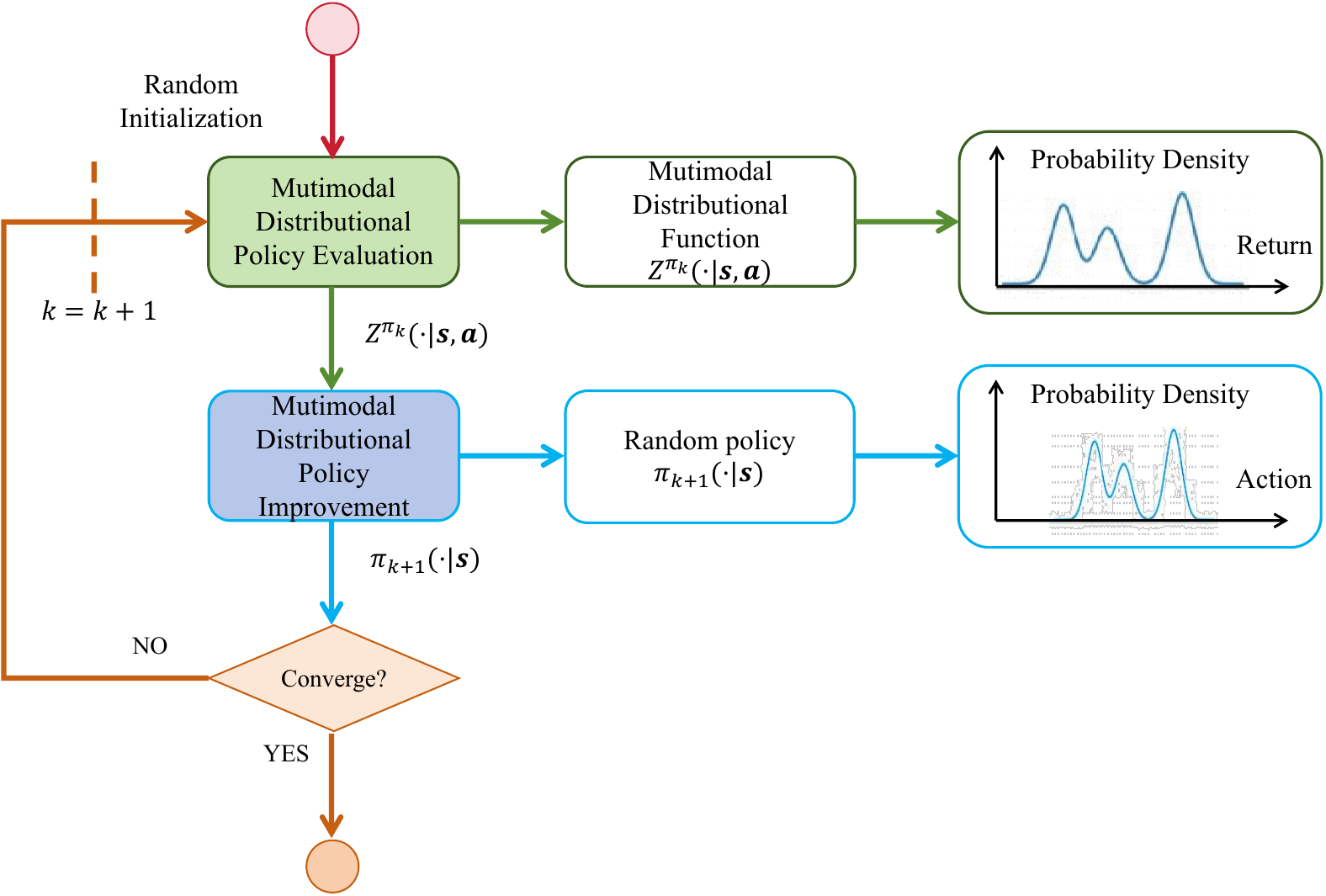}
\caption{Multimodal Distributional Policy Iteration Framework}
\label{fig:MDPI}
\end{figure}

By alternately performing the multimodal distributional policy evaluation and improvement steps, we can prove that for all state-action pairs \((s, a)\), the multimodal Q-value \(Q^{\pi_{k}}(s, a)\) corresponding to the policy \(\pi_{k}\) strictly monotonically increases with the number of iterations \(k\). This process is called Multimodal Distributional Policy Iteration (MDPI), and its framework is shown in Figure \ref{fig:MDPI}. 

\subsection{Diffusion Value Network}
For value distribution learning, the reverse diffusion process  of the diffusion model is regarded as a process of recovering the distribution of real return samples from noise.

If the diffusion model can characterize the complete multimodal distribution, the two distributions $P(Z(s,a))$ and $P(r(s,a)+Z(s',a'))$ must be close and converge. When constructing the diffusion value network, what is generated by reverse denoising is $Z^{\pi}(s, a)$. At this time, the return of the diffusion denoising process is denoted as $z_T$. In the reverse denoising process of the diffusion model, the given initial noise is \(z_T \sim \mathcal{N}(0, \mathbf{I})\), where $T$ is the total number of steps in the diffusion process.


Each transition in the reverse process is modeled as a Gaussian distribution. The posterior estimated distribution after removing the Gaussian noise added to $z_{t-1}$ is:
\begin{equation}
p_{\theta}(z_{t-1}\mid z_t)=\mathcal{N}\left(z_{t-1};\mu_{\theta}(z_t,t),\sum_{\theta}(z_t,t)\right),
\end{equation}
where $\mu_{\theta}\left({z}_t, t\right)$ and $\Sigma_{\theta}\left({z}_t, t\right)$ are learned functions parameterized by a deep neural network, which predict the mean and variance of the denoised estimate at each step.

The reverse denoising of the diffusion model generates the sample $z^0$ from $t = T, T-1, \dots, 0$, that is:
\begin{equation}
z_{t-1}=\frac{1}{\sqrt{\alpha_t}}\left(z_t-\frac{\beta_t}{\sqrt{1-\bar{\alpha}_t}}\epsilon_{\theta}(z_t, s, a, t)\right)+\sqrt{\beta_t}\epsilon,
\end{equation}
where $\epsilon \sim \mathcal{N}\left(0, \mathbf{I}\right)$ is an additional Gaussian noise used to introduce randomness in the denoising process; $\alpha_t = 1-\beta_t$, and $\beta_t$ is a predefined noise scheduling parameter that controls the degree of denoising at each step; $\bar{\alpha}_t = \prod_{k = 1}^t \alpha_k$ is the cumulative $\alpha$ value; $\epsilon_{\theta}$ is a parameterized noise prediction network, and $\theta$ is the parameter of the network, which predicts the noise according to the current noise sample $z_t$, state $s$, action $a$ and the number of steps $t$.


At this time, based on the predicted $\mu_{\theta}$ and the true $\tilde{\mu}$, $L_t$ is used as the loss function (optimization objective) in the training process to minimize the difference:
\begin{equation}
\begin{aligned}
L_t&=\mathbb{E}_{z_0,\epsilon}\left[\frac{1}{2\left\|\Sigma_{\theta}(z_t,t)\right\|^2}\left\|\tilde{\mu}_t(z_t,z_0)-\mu_{\theta}(z_t,t)\right\|^2\right].
\end{aligned}
\end{equation}

When training the diffusion model, ignoring the weight term to simplify $L_t$ has a better effect. Therefore, the above formula can be simplified as:
\begin{equation}
\begin{aligned}
L_t^{\text{simple}}&=\mathbb{E}_{t\sim[1,T],z_0,\epsilon_t}\left[\left\|\epsilon_t-\epsilon_{\theta}\left(\sqrt{\bar{\alpha}_t}z_0+\sqrt{1-\bar{\alpha}_t}\epsilon_t,t\right)\right\|^2\right].
\end{aligned}
\end{equation}

That is, the goal of model training is to minimize the Kullback-Leibler (KL) divergence between the approximate inverse process and the true posterior, which is used to measure the difference between the true posterior distribution $q({z}_{t-1} \mid {z}_t, {z}_0)$ and the model approximation $p_{\theta}({z}_{t-1} \mid {z}_t)$. 

\subsection{Distributional Soft Actor-Critic with Diffusion Policy} 
The distribution of the diffusion policy lacks an analytical expression, so its entropy cannot be directly determined. However, in the same state, multiple samples can be used to obtain a series of actions. By fitting these action points, the action distribution corresponding to that state can be estimated. In this paper, a Gaussian Mixture Model (GMM) is used to fit the policy distribution. The GMM forms a complex probability density function by combining multiple Gaussian distributions and can be expressed as:
\begin{equation}
\label{eq:GMM-3}
\hat{f}(a)=\sum_{k = 1}^{K}w_k\cdot\mathcal{N}\left(a\mid\mu_k,\Sigma_k\right),
\end{equation}
where $K$ is the number of Gaussian distributions, $w_k$ is the mixing weight of the $k$-th component, satisfying $\sum_{k = 1}^K w_k = 1$ and $w_k \geq 0$.

For each state, $N$ actions $a_1, a_2, \ldots, a_N \in A$ are sampled using the diffusion policy. Then the parameters of the GMM are estimated using the Expectation-Maximization (EM) algorithm. In the expectation step, the posterior probability that each data point $a_i$ belongs to each component $k$ is calculated and expressed as:
\begin{equation}
\gamma(z_{ik})=\frac{w_k\cdot\mathcal{N}\left(a_i\mid\mu_k,\Sigma_k\right)}{\sum_{j = 1}^{K}w_j\cdot\mathcal{N}\left(a_i\mid\mu_j,\Sigma_j\right)},
\end{equation}
where $\gamma(z_{ik})$ represents the probability that the observed data $a_i$ comes from the $k$-th component under the current parameter estimation.

According to Equation \eqref{eq:GMM-3}, the entropy of the action distribution corresponding to the state can be estimated in the following way:
\begin{equation}
\mathcal{H}_{{s}}  = -\sum_{k = 1}^{K}w_k\log\left(w_k\right)+\sum_{k = 1}^{K}w_k\cdot\frac{1}{2}\log\left((2\pi e)^d \mid \Sigma_k \mid \right),
\end{equation}
where $d$ is the dimension of the action. Then the average value of the action entropies of a batch of states is selected as the estimated entropy $\hat{\mathcal{H}}$ of the diffusion policy.

To enable the algorithm to adaptively determine the value of the policy entropy coefficient $\alpha$, the mechanism of adaptive policy entropy adjustment is selected, and its specific update method is:
\begin{equation}
\alpha\leftarrow\alpha-\beta_{\alpha}\left(\hat{\mathcal{H}}-\bar{\mathcal{H}}\right),
\end{equation}
where $\beta_\alpha$ represents the learning rate of the entropy coefficient, $\hat{\mathcal{H}}$ represents the expected value of the policy entropy, $\bar{\mathcal{H}}$ represents the target value of the policy entropy, and usually $\bar{\mathcal{H}} = -\mathrm{dim}(A)$.

As the algorithm learning progresses, the value of the policy entropy coefficient $\alpha$ will gradually decrease, so that the expected value $\hat{\mathcal{H}}$ of the policy entropy in each state approaches the target value $\bar{\mathcal{H}}$ of the policy entropy, thus ultimately ensuring the convergence performance of the algorithm policy.

\begin{figure}
\centering
\includegraphics[width=0.45\textwidth]{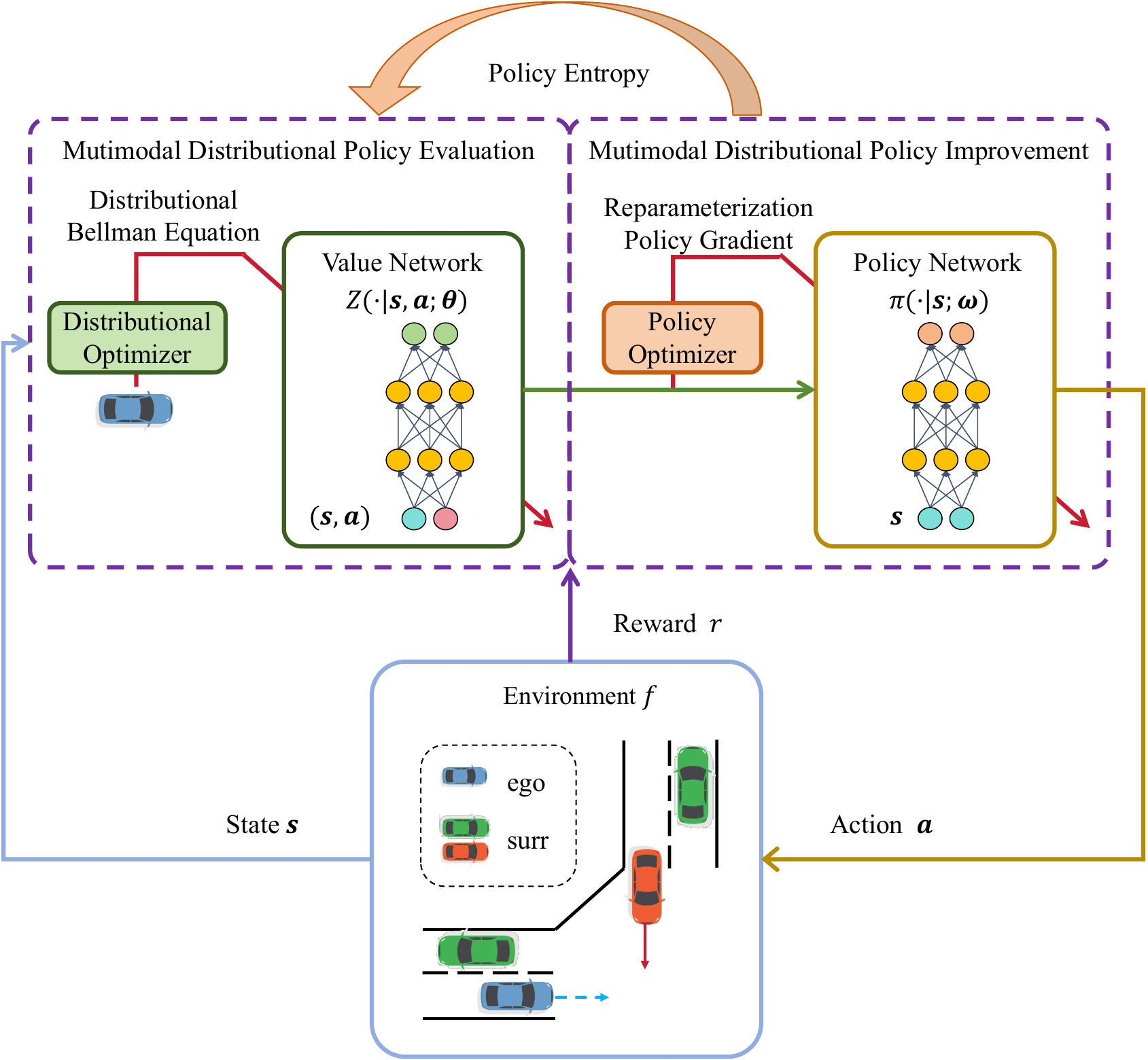}
\caption{DSAC-D Algorithm Framework}
\label{fig:DSAC-D}
\end{figure}

According to the above content, the DSAC-D (Distributional Soft Actor-Critic with Diffusion Policy) reinforcement learning algorithm can be derived, and its complete pseudocode is shown in Algorithm \ref{alg:DSAC-D}. The algorithm flowchart can be seen in the Figure \ref{fig:DSAC-D}.

\begin{algorithm}
  \caption{DSAC-D Algorithm}
  \label{alg:DSAC-D}
  \small
  \begin{algorithmic}
    \STATE Initialize the parameters $\theta$ of the value distribution network, the parameters $\omega$ of the policy network, and the policy entropy coefficient $\alpha$.
    \STATE Initialize the parameters of the target network: $\theta' \leftarrow \theta$ and $\omega' \leftarrow \omega$.
    \STATE Select appropriate learning rates $\beta_{z}$, $\beta_{\pi}$, $\beta_{\alpha}$, and $\tau$.
    \STATE Set the initial iteration step number $k = 0$.
    \FOR{Each iteration}
      \FOR{Each sampling step}
        \STATE Sample $a \sim \pi_{\theta}(\cdot \mid s)$.
        \STATE Add noise: $a = a + \lambda\alpha \cdot \mathcal{N}\left(0, \mathbf{I}\right)$.
        \STATE Obtain the reward $r$ and the new state $s^{\prime}$.
        \STATE Store the data tuple $(s, a, r, s^{\prime})$ in the replay buffer $\mathcal{B}$.
      \ENDFOR
      \FOR{Each update step}
        \STATE Sample data $(s, a, r, s^{\prime})$ from $\mathcal{B}$.
        \STATE Calculate and update the diffusion value network: $\theta \leftarrow \theta-\beta_{z} \nabla_{\theta} J_{z}(\theta)$.
        \IF{The iteration step number $k$ is divisible by $10000$}
          \STATE Update the diffusion policy network: $\omega \leftarrow \omega + \beta_{\pi} \nabla_{\omega} J_{\pi}(\omega)$.
          \STATE Estimate the entropy of the diffusion policy: $\hat{\mathcal{H}} = \mathbb{E}_{s \sim \mathcal{B}} [\mathcal{H}_s]$.
          \STATE Update the policy entropy coefficient: $\alpha\leftarrow\alpha-\beta_{\alpha}\left(\hat{\mathcal{H}}-\bar{\mathcal{H}}\right)$.
          \STATE Update the target value network: $\theta' \leftarrow \tau\theta + (1-\tau)\theta'$.
          \STATE Update the target policy network: $\omega' \leftarrow \tau\omega + (1-\tau)\omega'$.
        \ENDIF
      \ENDFOR
    \ENDFOR
  \end{algorithmic}
\end{algorithm}

\section{Experiments}

\subsection{Environments}
\label{sec:compare Mujoco}

\begin{figure} 
    \centering   
    \begin{subfigure}[t]{0.15\textwidth}    
        \includegraphics[width=\textwidth]{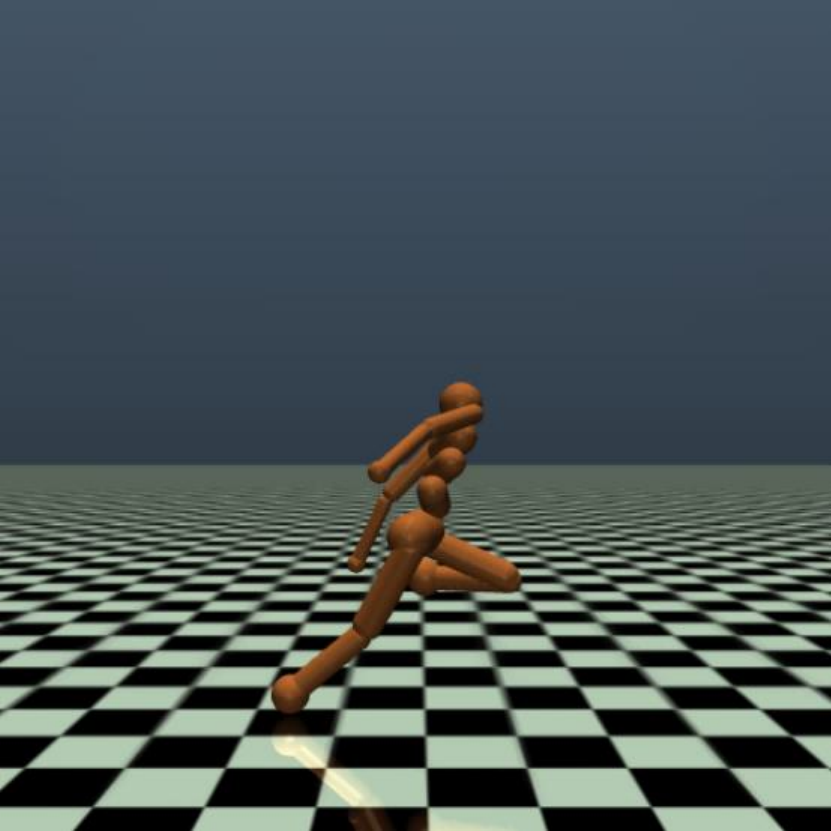}    
        \caption*{(a) Humanoid-v3}   
    \end{subfigure}   
    \begin{subfigure}[t]{0.15\textwidth}    
        \includegraphics[width=\textwidth]{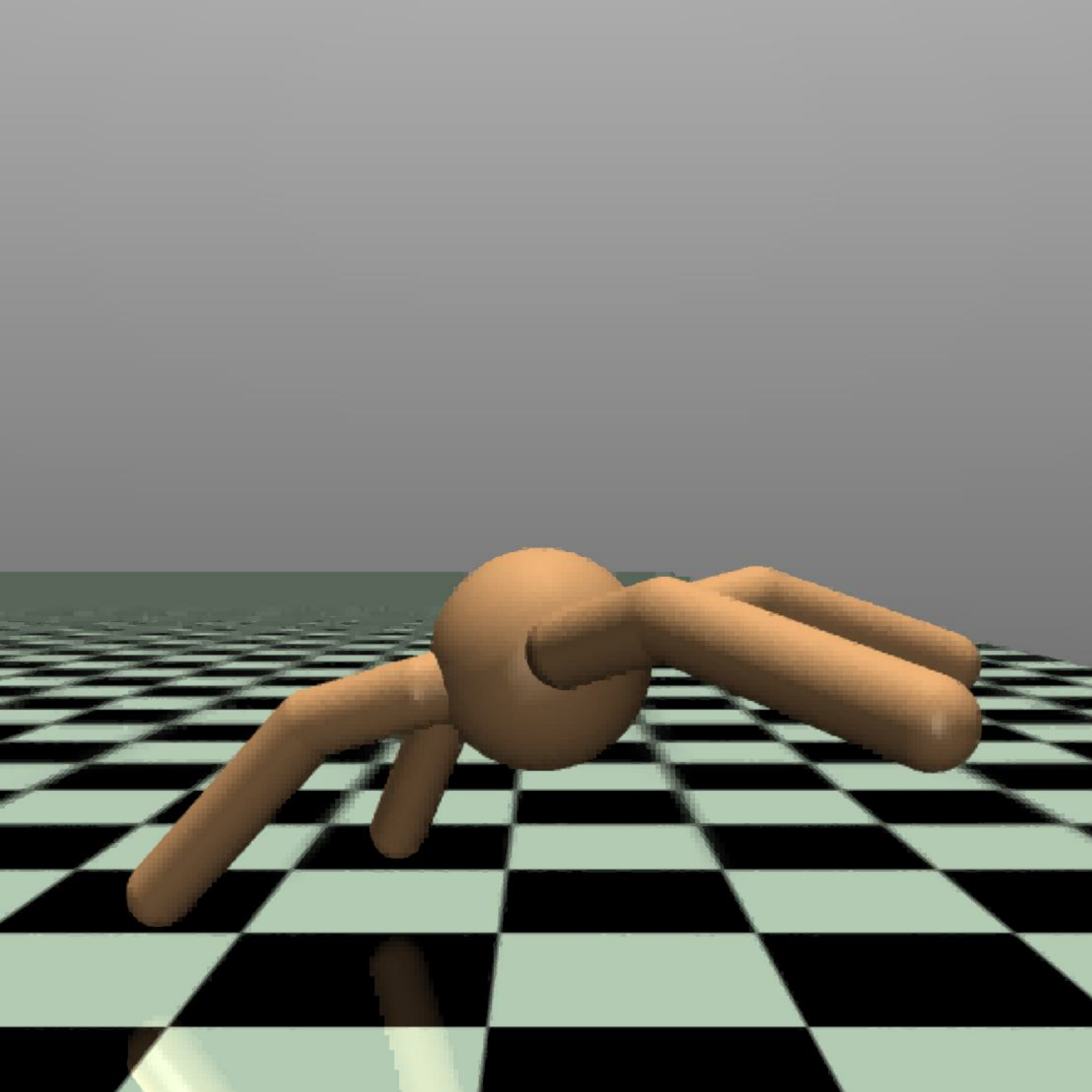}     
        \caption*{(b) Ant-v3}   
    \end{subfigure}   
    \begin{subfigure}[t]{0.15\textwidth}    
        \includegraphics[width=\textwidth]{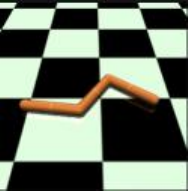}     
        \caption*{(c) Swimmer-v3}   
    \end{subfigure}   
    \vspace{0.5em} 
    \begin{subfigure}[t]{0.15\textwidth}    
        \includegraphics[width=\textwidth]{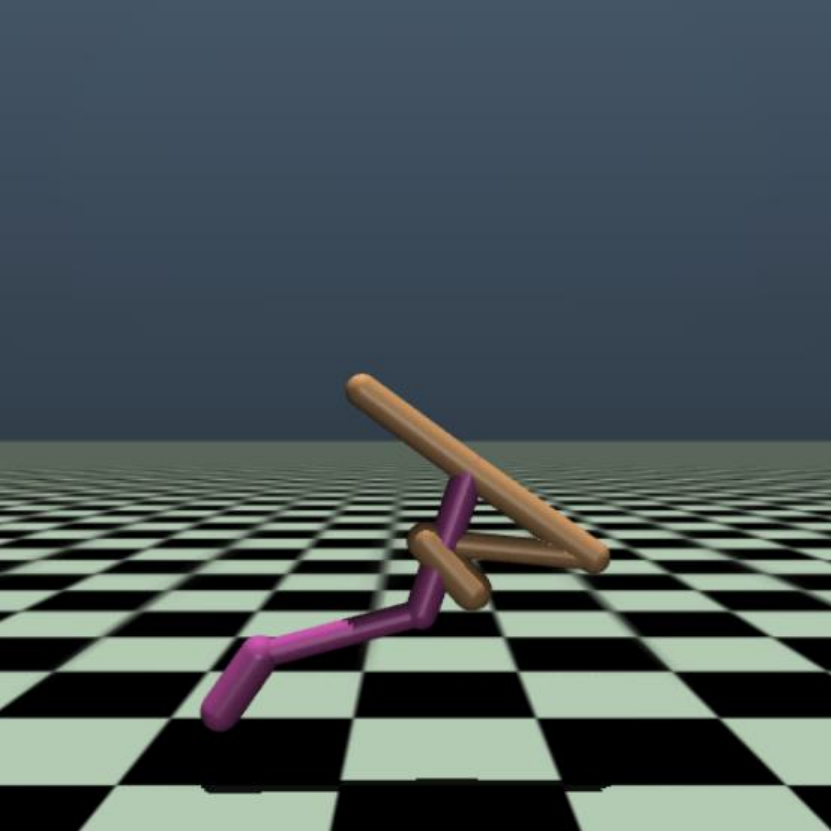}    
        \caption*{(d) Walker2d-v3}   
    \end{subfigure}   
    \begin{subfigure}[t]{0.15\textwidth}    
        \includegraphics[width=\textwidth]{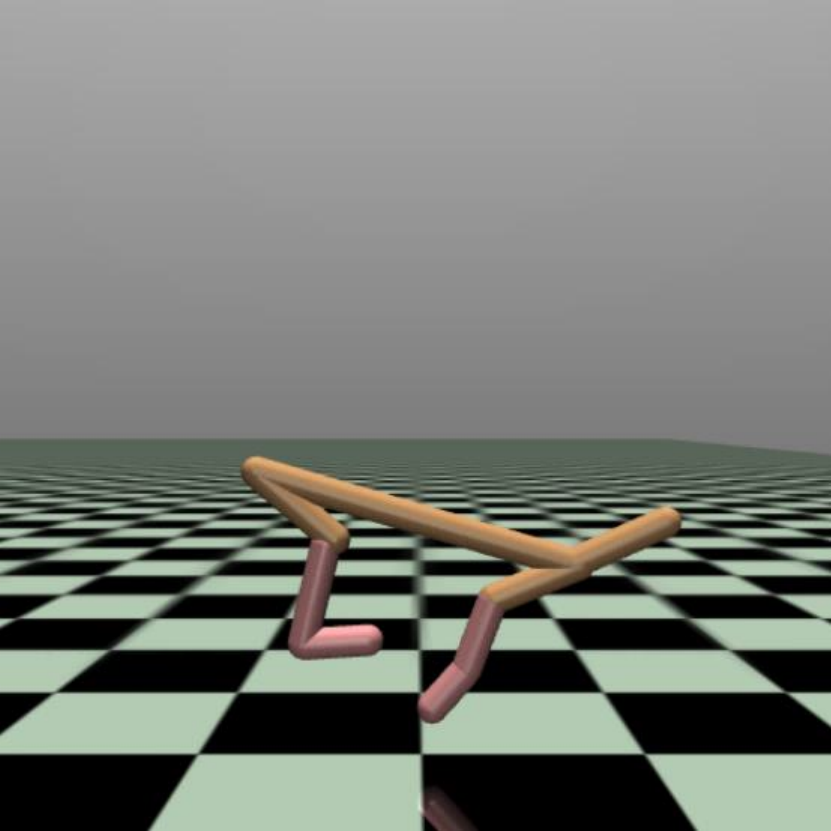} 
        \caption*{(e) HalfCheetah-v3}   
    \end{subfigure}   
    \begin{subfigure}[t]{0.15\textwidth}    
        \includegraphics[width=\textwidth]{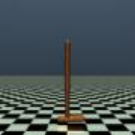}     
        \caption*{(f) Hopper-v3}   
    \end{subfigure}   
    \vspace{0.5em} 
    \begin{subfigure}[t]{0.15\textwidth}    
        \includegraphics[width=\textwidth]{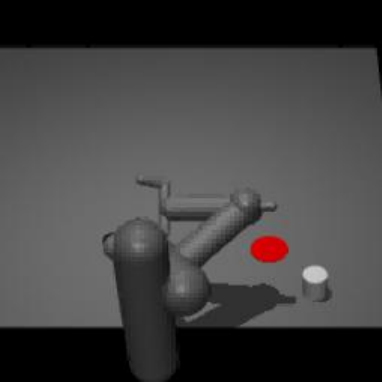}   
        \caption*{(g) Pusher-v2}   
    \end{subfigure}   
    \begin{subfigure}[t]{0.15\textwidth}    
        \includegraphics[width=\textwidth]{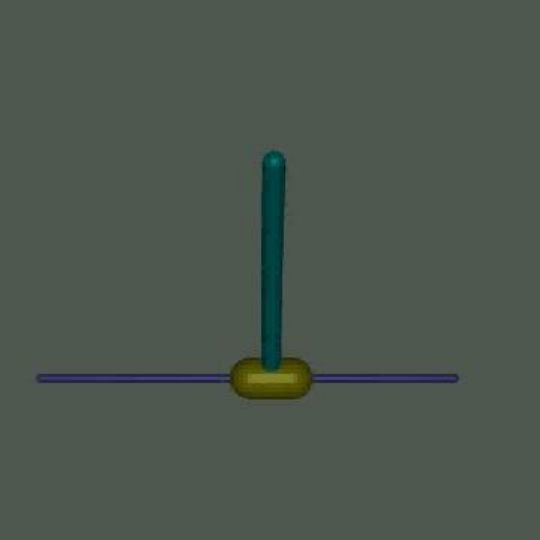} 
        \caption*{(h) IDP-v3}   
    \end{subfigure}   
    \begin{subfigure}[t]{0.15\textwidth}    
        \includegraphics[width=\textwidth]{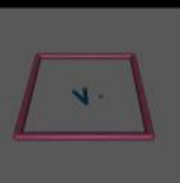}   
        \caption*{(i) Reacher-v2}   
    \end{subfigure}   
    \caption{MuJuCo simulation task environment}   
    \label{fig:mujoco} 
\end{figure}

\textbf{MuJoCo}: This widely-used benchmark simulates multi-joint robotic systems~\cite{tassa2018deepmind}. In this study, we focus on 9 challenging tasks: Humanoid-v3, Ant-v3, HalfCheetah-v3, Walker2d-v3, InvertedDoublePendulum-v3, Hopper-v3, Pusher-v2, Reacher-v2 and Swimmer-v3, as depicted in Figure~\ref{fig:mujoco}. 

\textbf{Vehicle Meeting}: We designed a vehicle trajectory tracking and obstacle avoidance environment with three-degree-of-freedom and perimeter vehicle constraints. In this intersection vehicle-meeting scenario, the surrounding vehicles' driving styles (aggressive, normal, conservative) are randomly selected at the start, resulting in different accelerations near the meeting point. The environment's reward function penalizes lateral and longitudinal deviations, speed, angular velocity, acceleration, etc. The ego-vehicle must learn multi-modal Q-values and policies in this stochastic setting. The training environment is based on GOPS\cite{wang2023gops} and IDSIM \cite{jiang2024reinforcement}. In this work, the diffusion policy is also integrated into GOPS as an approximate function module, which is referred to as DiffusionNet. 


\subsection{Algorithm Performance and Multimodal Policy}

In algorithm performance and policy representation experiments, We compare DSAC-D with several well-established online RL algorithms, including DACER~\cite{wang2024diffusion, wang2025enhanced}, DDPG~\cite{lillicrap2016continuous}, TD3~\cite{fujimoto2018addressing}, PPO~\cite{schulman2017proximal}, SAC~\cite{haarnoja2018soft}, DSACT~\cite{Duan2025dsact}, and TRPO \cite{schulman2015trust}. These baselines, available in the GOPS solver, have been widely tested. For fair comparisons, we run 20 parallel environment interactions per iteration. All algorithms use a three-layer MLP with GeLU \cite{hendrycks2016gaussian} or Mish \cite{misra2019mish} and Adam \cite{kingma2014adam} for optimization.

\begin{figure*}[ht!]
\centering   
    \begin{subfigure}[t]{0.3\textwidth}     \includegraphics[width=\textwidth]{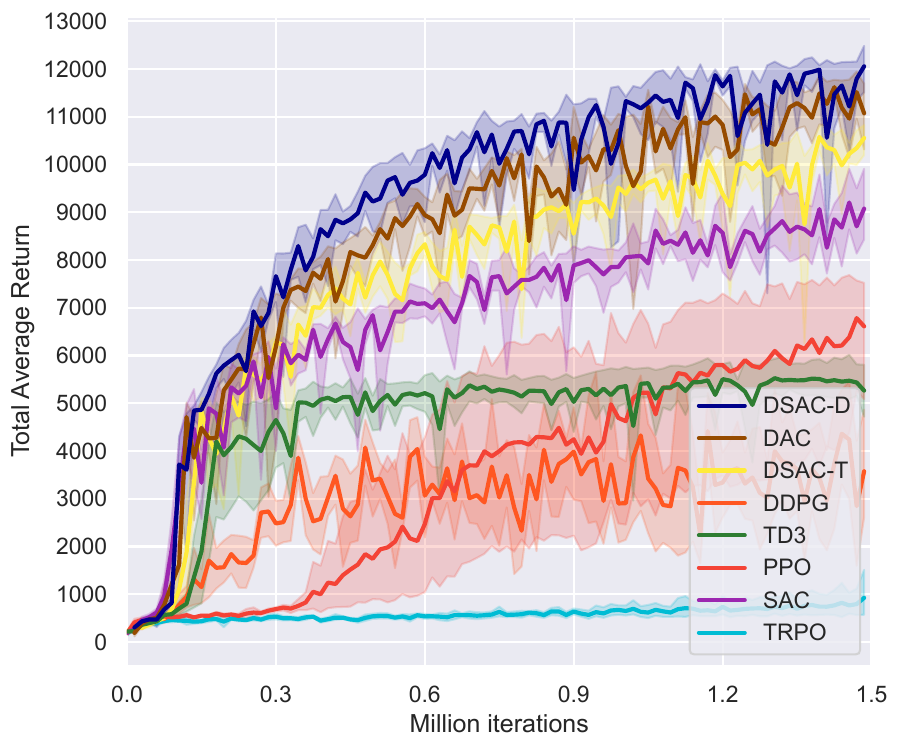}     
        \caption*{(a) Humanoid-v3}   
    \end{subfigure}   
    \begin{subfigure}[t]{0.3\textwidth}     \includegraphics[width=\textwidth]{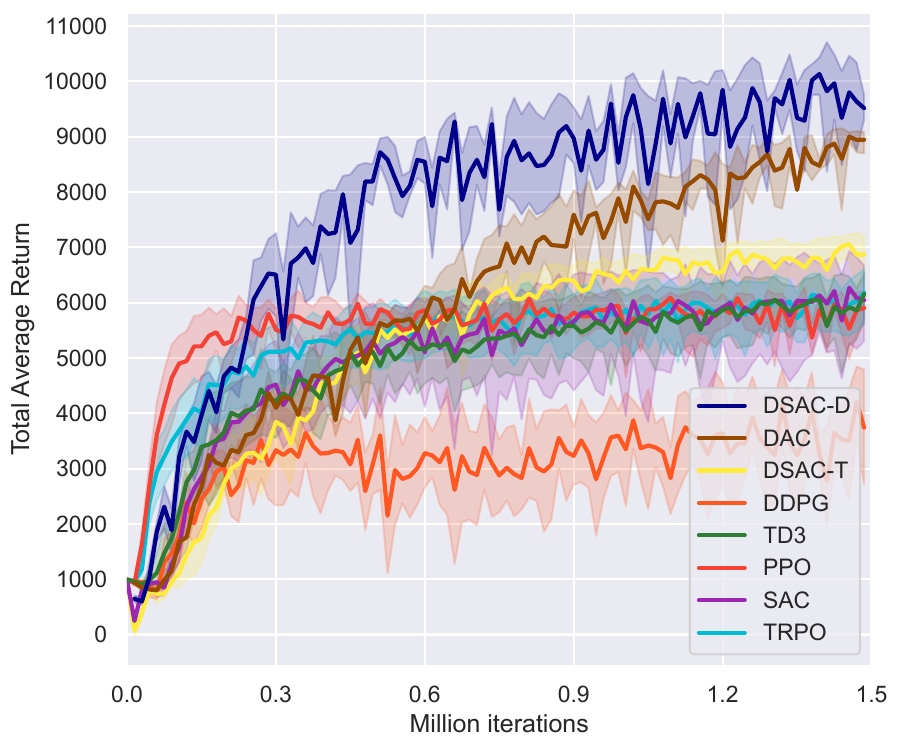} 
        \caption*{(b) Ant-v3}   
    \end{subfigure}   
    \begin{subfigure}[t]{0.3\textwidth}   \includegraphics[width=\textwidth]{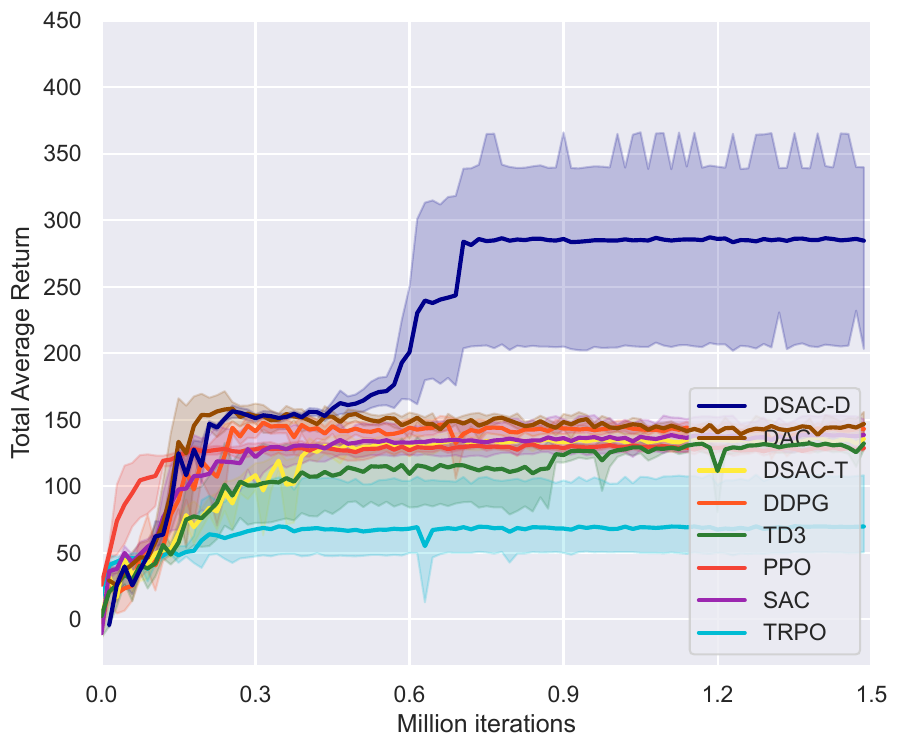}
        \caption*{(c) Swimmer-v3}   
    \end{subfigure}   
    \vspace{0.5em} 
    \begin{subfigure}[t]{0.3\textwidth} \includegraphics[width=\textwidth]{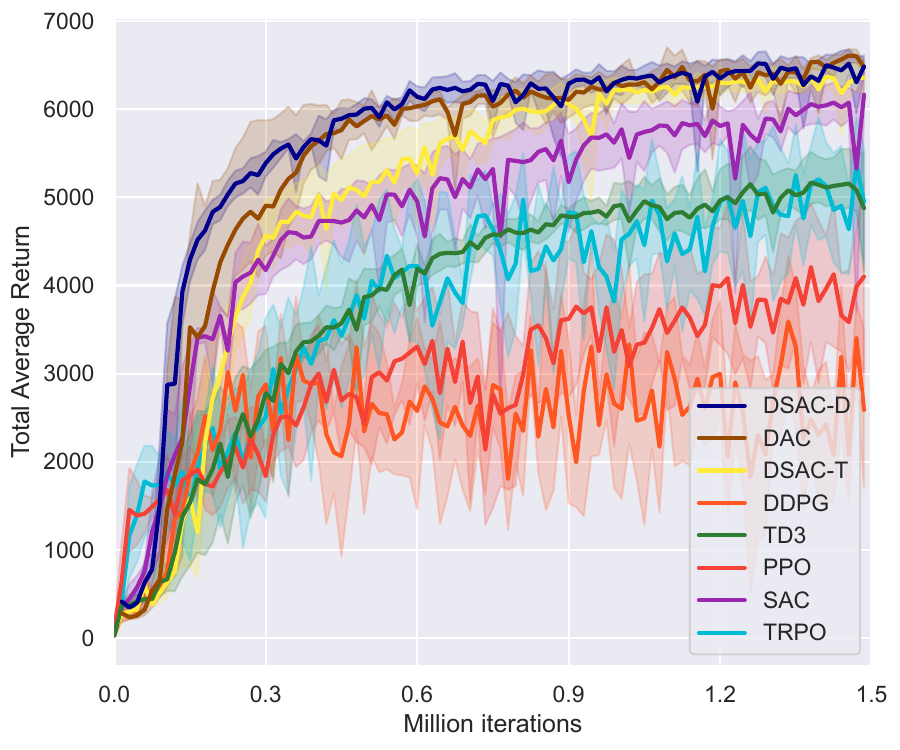}
        \caption*{(d) Walker2d-v3}   
    \end{subfigure}   
    \begin{subfigure}[t]{0.3\textwidth}
    \includegraphics[width=\textwidth]{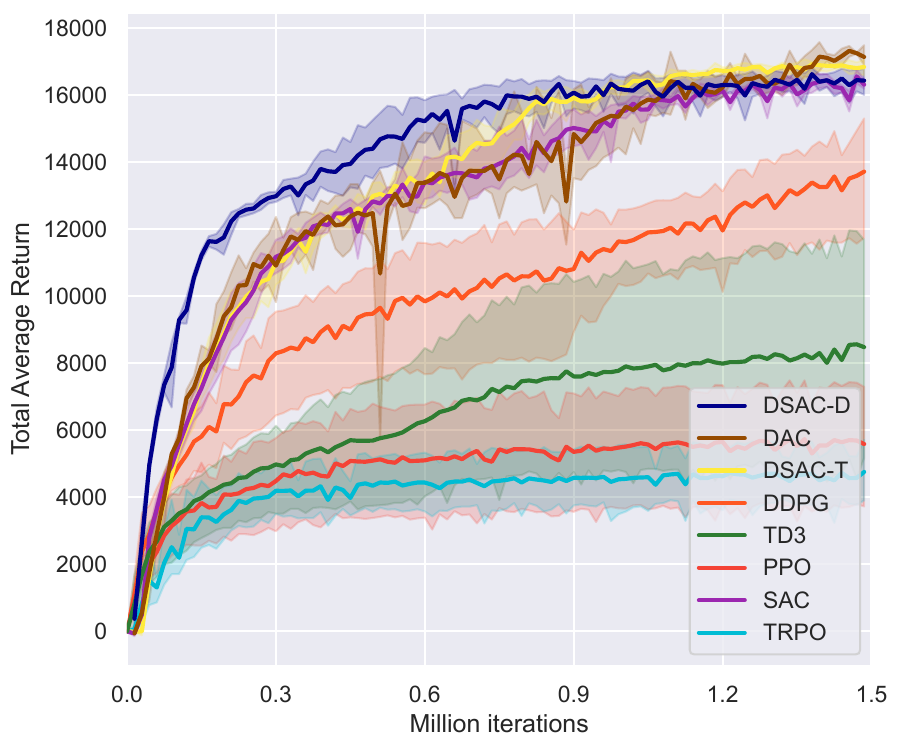} 
        \caption*{(e) HalfCheetah-v3} 
    \end{subfigure}   
    \begin{subfigure}[t]{0.3\textwidth}
    \includegraphics[width=\textwidth]{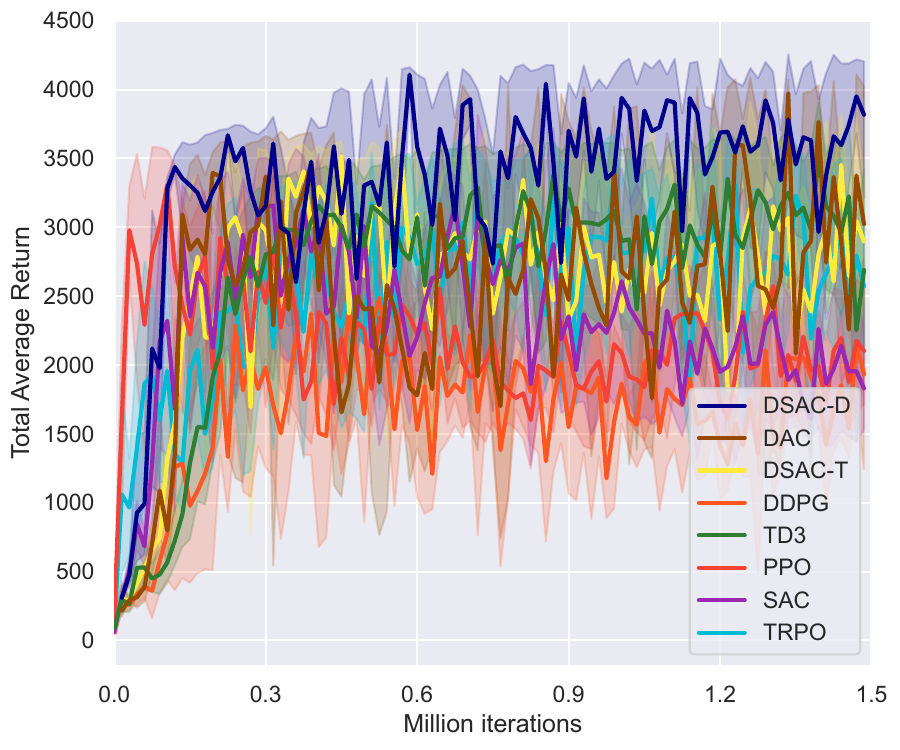}     
        \caption*{(f) Hopper-v3}   
    \end{subfigure}   
    \vspace{0.5em} 
    \begin{subfigure}[t]{0.3\textwidth}\includegraphics[width=\textwidth]{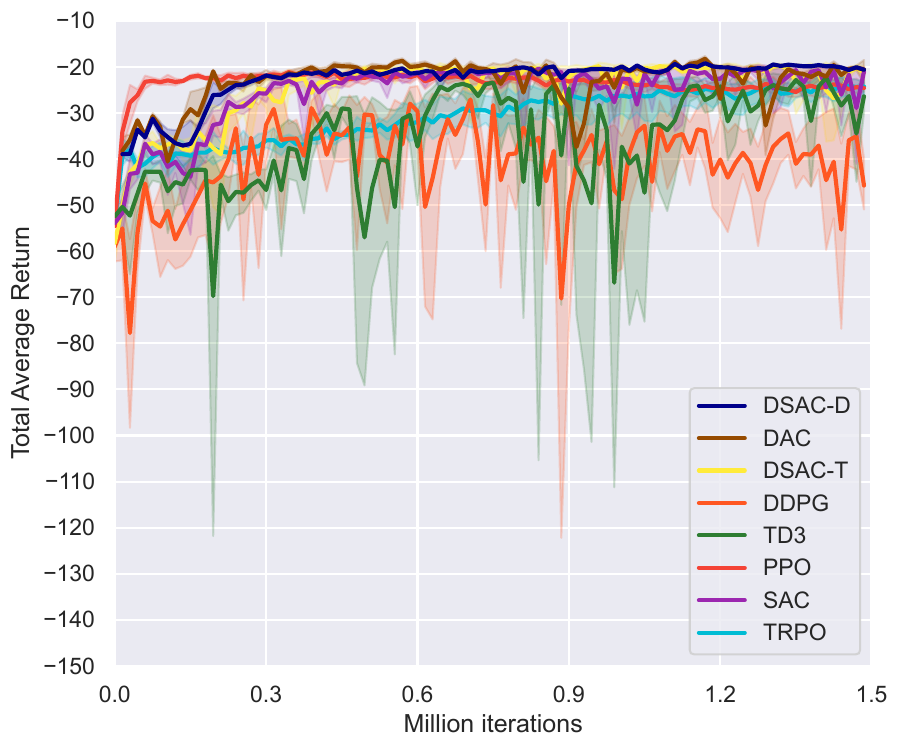} 
        \caption*{(g) Pusher-v2}   
    \end{subfigure}   
    \begin{subfigure}[t]{0.3\textwidth}     \includegraphics[width=\textwidth]{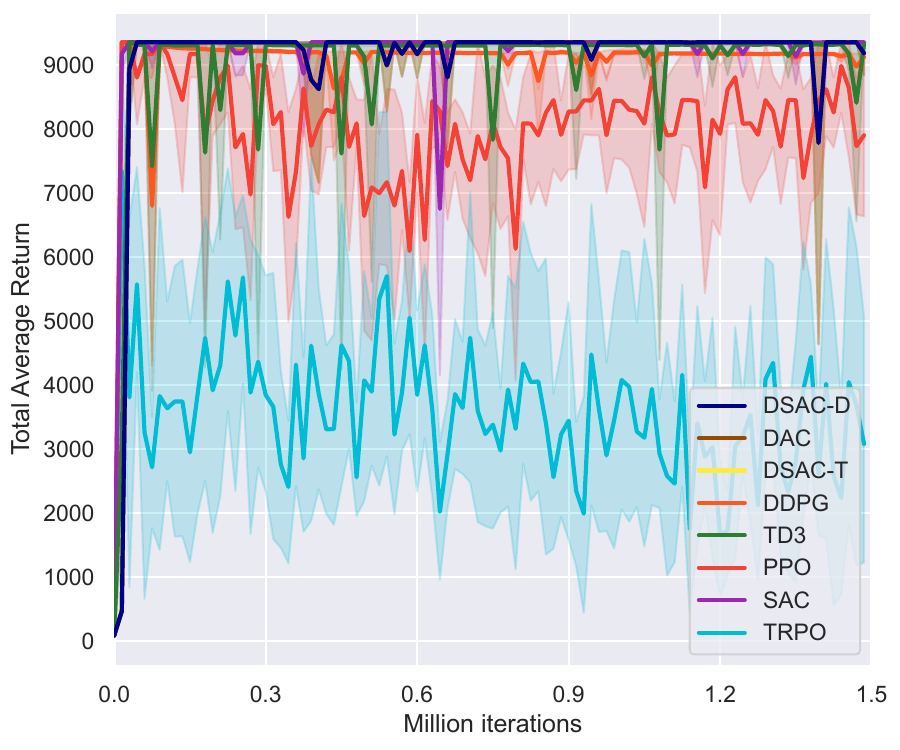} 
        \caption*{(h) InvertedDoublePendulum-v3}   
    \end{subfigure}   
    \begin{subfigure}[t]{0.3\textwidth}     \includegraphics[width=\textwidth]{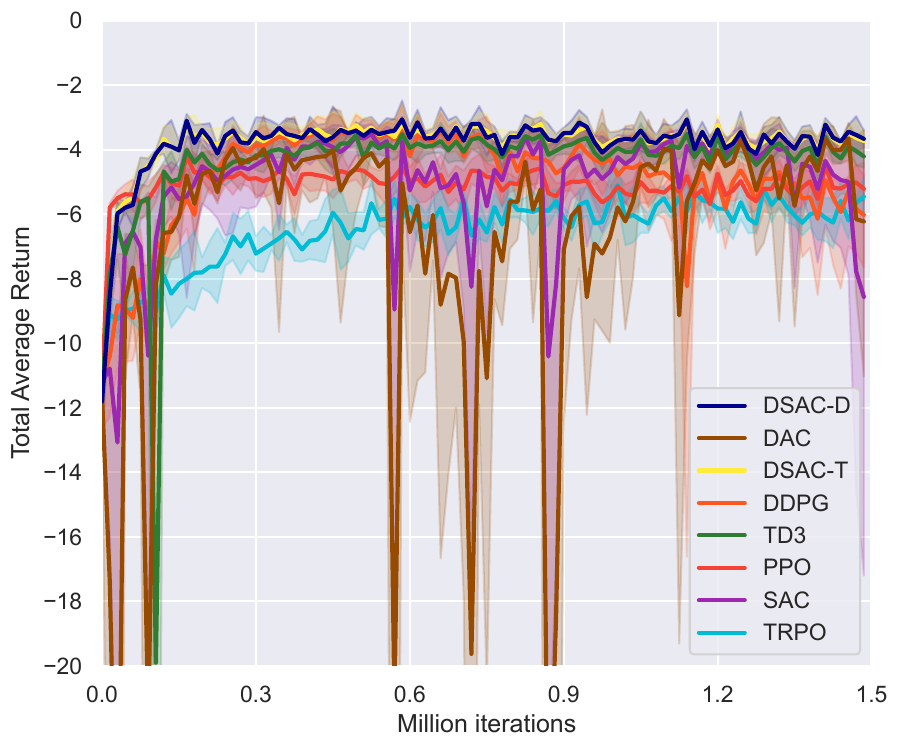}   
        \caption*{(i) Reacher-v2}   
    \end{subfigure} 
\caption{\textbf{Training curves on benchmarks.} The solid lines represent the mean, while the shaded regions indicate the 95\% confidence interval over five runs. The iteration of PPO and TRPO is measured by the number of network updates.}
\label{fig:benchmark}
\end{figure*}

Using MuJoCo benchmarks, we conduct 9 independent tests for each algorithm to assess DSAC-D's robotic control performance. The results in Figure~\ref{fig:benchmark} and Table~\ref{tab:benchmark} show DSAC-D outperforms all baselines in all tasks. It converges quickly in various environments, leveraging multi-modal distribution learning to avoid local optima. 

In the Ant-v3 task, DSAC-D achieves relative improvements of 6.3\%, 36.7\%, 50.7\%, 56.6\%, 112.9\%, 56.1\%, and 57.3\% compared to DAC, DSAC-T, SAC, TD3, DDPG, TRPO, and PPO, respectively. In the Swimmer-v3 task, DSAC-D achieves relative improvements of 87.1\%, 106.0\%, 104.3\%, 113.4\%, 95.9\%, 308.6\%, and 119.9\% compared to DAC, DSAC-T, SAC, TD3, DDPG, TRPO, and PPO, respectively.

\begin{table*}[!ht]
 \centering

    \caption{\textcolor{black}{\textbf{Comparison of Algorithm Performance.} Computed as the mean of the highest return values observed in the final 10\% of iteration steps per run, with an evaluation interval of 15,000 iterations. The maximum value for each task is bolded. $\pm$ corresponds to standard deviation over five runs.}}

	\label{tab:benchmark}
   \resizebox{\textwidth}{!}{ 
\begin{tabular}{c c c c c c c c c}
  \toprule
  Task & DSAC-D & DAC & DSAC-T & SAC & TD3 & DDPG & TRPO & PPO \\ \hline
  Humanoid-v3 & \textbf{11604} $\pm$ \textbf{791} & 11257 $\pm$ 608& 10829 $\pm$ 243 & 9335 $\pm$ 695 & 5631 $\pm$ 435 & 5291 $\pm$ 662 & 965 $\pm$ 555 & 6869 $\pm$ 1563 \\
  Ant-v3 & \textbf{9684} $\pm$ \textbf{882}& 9108 $\pm$ 103& 7086 $\pm$ 261& 6427 $\pm$ 804 & 6184 $\pm$ 486 & 4549 $\pm$ 788 & 6203 $\pm$ 578 & 6156 $\pm$ 185 \\
  Swimmer-v3 & \textbf{286} $\pm$ \textbf{104} & 152 $\pm$ 7 &  138$\pm$6& 140 $\pm$ 14 & 134 $\pm$ 5 & 146 $\pm$ 4 & 70 $\pm$ 38& 130 $\pm$ 2 \\
  InvertedDoublePendulum-v3 & \textbf{9360} $\pm$ \textbf{0}& \textbf{9360} $\pm$ \textbf{0} & \textbf{9360} $\pm$ \textbf{0} & \textbf{9360} $\pm$ \textbf{0} & 9347 $\pm$ 15 & 9183 $\pm$ 9 & 6259 $\pm$ 2065 & 9356 $\pm$ 2 \\
  Hopper-v3 & \textbf{4573} $\pm$ \textbf{203}& 4104 $\pm$ 49&  3660$\pm$533& 2483 $\pm$ 943 & 3569 $\pm$ 455 & 2644 $\pm$ 659 & 3474 $\pm$ 400 & 2647 $\pm$ 482 \\
    Pusher-v2 & \textbf{-19} $\pm$ \textbf{0} & 19 $\pm$ 1 &  -19 $\pm$ 1 & -20 $\pm$ 0 & -21 $\pm$ 1 & -30 $\pm$ 6 & -23 $\pm$ 2 & -23 $\pm$ 1 \\
    HalfCheetah-v3 & \textbf{16409} $\pm$ \textbf{477}& 17177 $\pm$ 176& 17025 $\pm$ 157 & 16573 $\pm$ 224 & 8632 $\pm$ 4041 & 13970 $\pm$ 2083 & 4785 $\pm$ 967 & 5789 $\pm$ 2200 \\
  Walker2d-v3 & \textbf{6732} $\pm$ \textbf{89}& 6701 $\pm$ 62& 6424 $\pm$ 147 & 6200 $\pm$ 263 & 5237 $\pm$ 335 & 4095 $\pm$ 68 & 5502 $\pm$ 593 & 4831 $\pm$ 637 \\
    Reacher-v2 & \textbf{-3} $\pm$ \textbf{0} & \textbf{-3} $\pm$ \textbf{0} &  \textbf{-3} $\pm$ \textbf{0} & \textbf{-3} $\pm$ \textbf{0} & \textbf{-3} $\pm$ \textbf{0} & -4 $\pm$ 1 & -5 $\pm$ 1 & -4 $\pm$ 0 \\
  \bottomrule
\end{tabular}
}
\end{table*}

The value function estimation bias is the difference between the value network's output and the true value function. In real-world control tasks, getting the true value directly is hard due to system complexity and environmental uncertainty. But by using the Q-value definition and continuous reward data from simulations, we can calculate an approximate true value. Table \ref{tab:relative_bias} shows the average relative estimation biases of DSAC-D and several mainstream RL algorithms across various tasks. This indicates that by learning the value distribution function, the DSAC-D algorithm can effectively suppress the bias caused by overestimation and greatly improve the estimation accuracy of values function.

\begin{table*}
  \centering
  \caption{\textbf{Average relative value estimation bias over five runs.} The relative bias is computed using (estimate Q-value\(-\)true Q-value), where true Q-value is assessed based on the discounted accumulation of sampled rewards. The best value is bolded. The superscript\(^\star\)indicates superior estimation accuracy of DSAC-D over off-policy baselines including SAC, TD3, and DDPG. Meanwhile,\(^\dagger\)denotes superior estimation accuracy of DSAC-D over on-policy baselines like TRPO and PPO. When biases are comparable, underestimation is more favorable than overestimation.}
  \label{tab:relative_bias}
  \resizebox{\textwidth}{!}{
    \begin{tabular}{ccccccccc}
      \toprule
      Task                     & DSAC-D                 & DSAC-T                      & SAC    & TD3     & DDPG    & TRPO    & PPO     \\
      \hline
      Humanoid-v3            & \textbf{-21.26$^{\star \dagger}$} & -42.29 & -81.69 & -226.05 & 48.80   & 18.72   & 17.28   \\
      Ant-v3                 & \textbf{-3.55$^{\star \dagger}$} & -10.55 & -25.31 & -327.33 & 89.36   & 13.36   & 8.37    \\
      HalfCheetah-v3         & 21.45$^{\dagger}$     & 23.95       & \textbf{-4.82}  & -341.37 & 128.54  & 603.82  & 95.20   \\
      Walker2d-v3            & \textbf{-0.54$^{\star \dagger}$}  & -0.79  & -5.49  & -60.05  & 31.81   & 5.90    & 1.80    \\
      InvertedDoublePendulum-v3 & 2.42$^{\star}$       & 2.60        & 5.68   & -560.99 & 57632.34 & 3.32    & \textbf{1.57}    \\
      Hopper-v3              & \textbf{-2.58$^{\star \dagger}$}  & -5.13 & -6.00  & -27.86  & 74.85   & 2.65    & 4.83    \\
      Pusher-v2              & \textbf{-5.24$^{\star \dagger}$}  & -6.83  & -7.02  & -10.76  & 1.19    & -10.35  & -8.68   \\
      Reacher-v2             & -3.85             & -5.46             & -5.44  & -10.13  & \textbf{-0.28}   & -6.87   & -4.99   \\
      Swimmer-v3             & 0.10             & 0.60             & \textbf{-0.07}  & -1.00   & 0.10    & 0.15    & 0.02    \\
      \bottomrule
    \end{tabular}
  }
\end{table*}

\begin{figure*}
\centering
  \begin{subfigure}{0.47\linewidth} 
    \includegraphics[width=\textwidth]{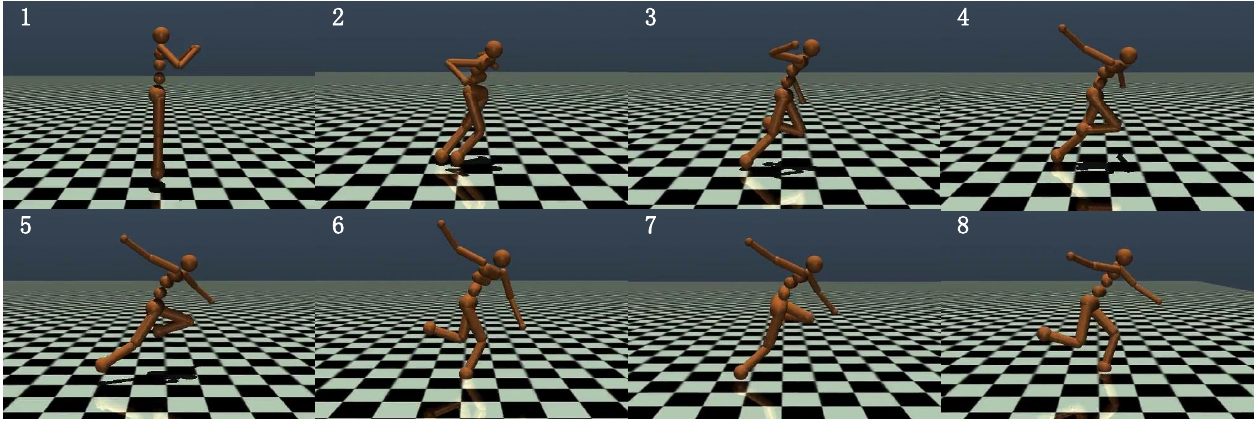} 
    \caption{Action mode 1 in Humanoid-v3 environment}
    \label{fig:Action Mode 1 in Humanoid-v3 Environment}
  \end{subfigure}
  \begin{subfigure}{0.47\linewidth}
    \includegraphics[width=\textwidth]{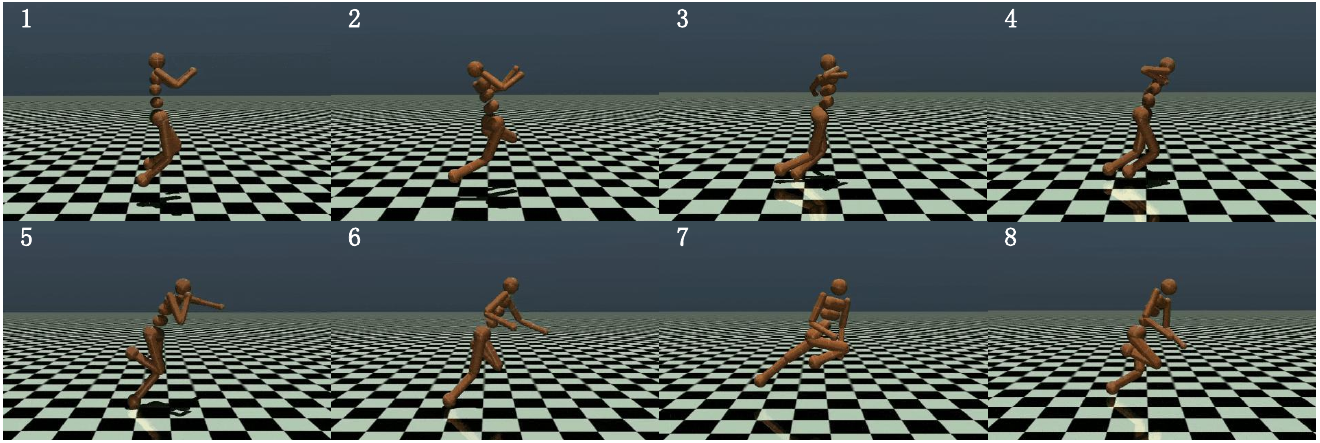} 
    \caption{Action mode 2 in Humanoid-v3 environment}
    \label{Action Mode 2 in Humanoid-v3 Environment}
  \end{subfigure}
  \begin{subfigure}{0.47\linewidth}
    \includegraphics[width=\textwidth]{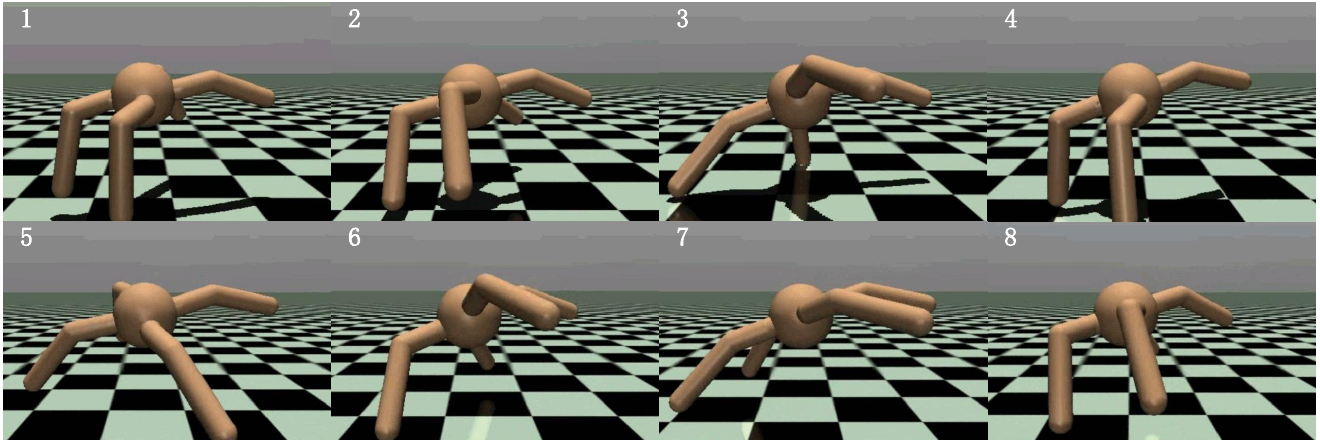} 
    \caption{Action mode 1 in Ant-v3 environment}
    \label{Action Mode 1 in Ant-v3 Environment}
    \end{subfigure}
  \begin{subfigure}{0.47\linewidth}
    \centering
    \includegraphics[width=\textwidth]{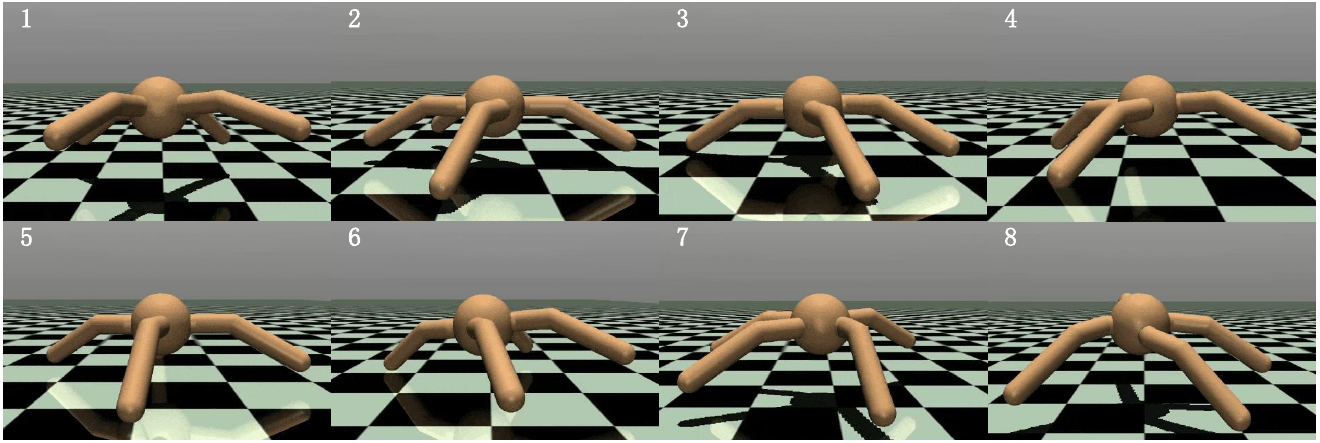} 
    \caption{Action mode 2 in Ant-v3 environment}
    \label{Action Mode 2 in Ant-v3 Environment}
  \end{subfigure}
  
\caption{Different decision-making action modes on MuJoCo tasks} 
\label{fig: multimodal_mujoco}
\end{figure*}

For the high-dimensional and complex control tasks Humanoid-v3 and Ant-v3, Figure~\ref{fig: multimodal_mujoco} shows the modalities of different action behaviors. In the Humanoid task, DSAC-D allows the agent to run with a human-like posture and natural arm swing, while DSAC-T's agent has a small stride, backward-leaning upper body, and unnatural arm movement. In the Ant-v3 task, DSAC-D shows policy multimodality, with both a four-legged, small-stride crawl and a three-legged, large-stride crawl, making its training curve rise quickly. In contrast, DSAC-T in the Ant-v3 task only has the single modality of a four-legged, small-stride crawl.  

\subsection{Vehicle Multimodal Trajectory}
\begin{figure*}[ht!]
    \centering
        \begin{subfigure}[b]{0.47\textwidth}
        \includegraphics[width=\textwidth]{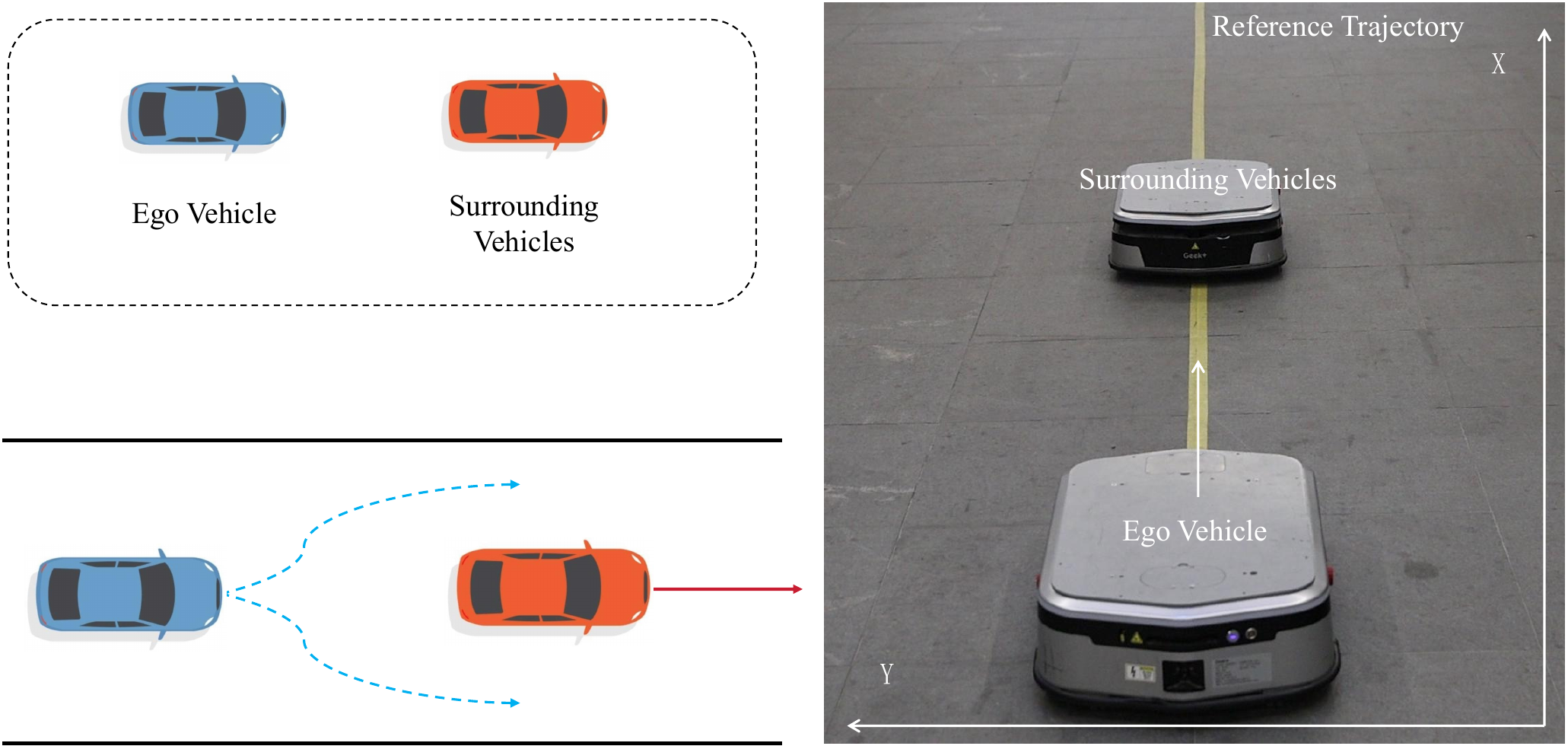}
        \caption{Experimental scenario 1}
        \label{Experimental Scene 1}
    \end{subfigure}
    \begin{subfigure}[b]{0.47\textwidth}
        \includegraphics[width=\textwidth]{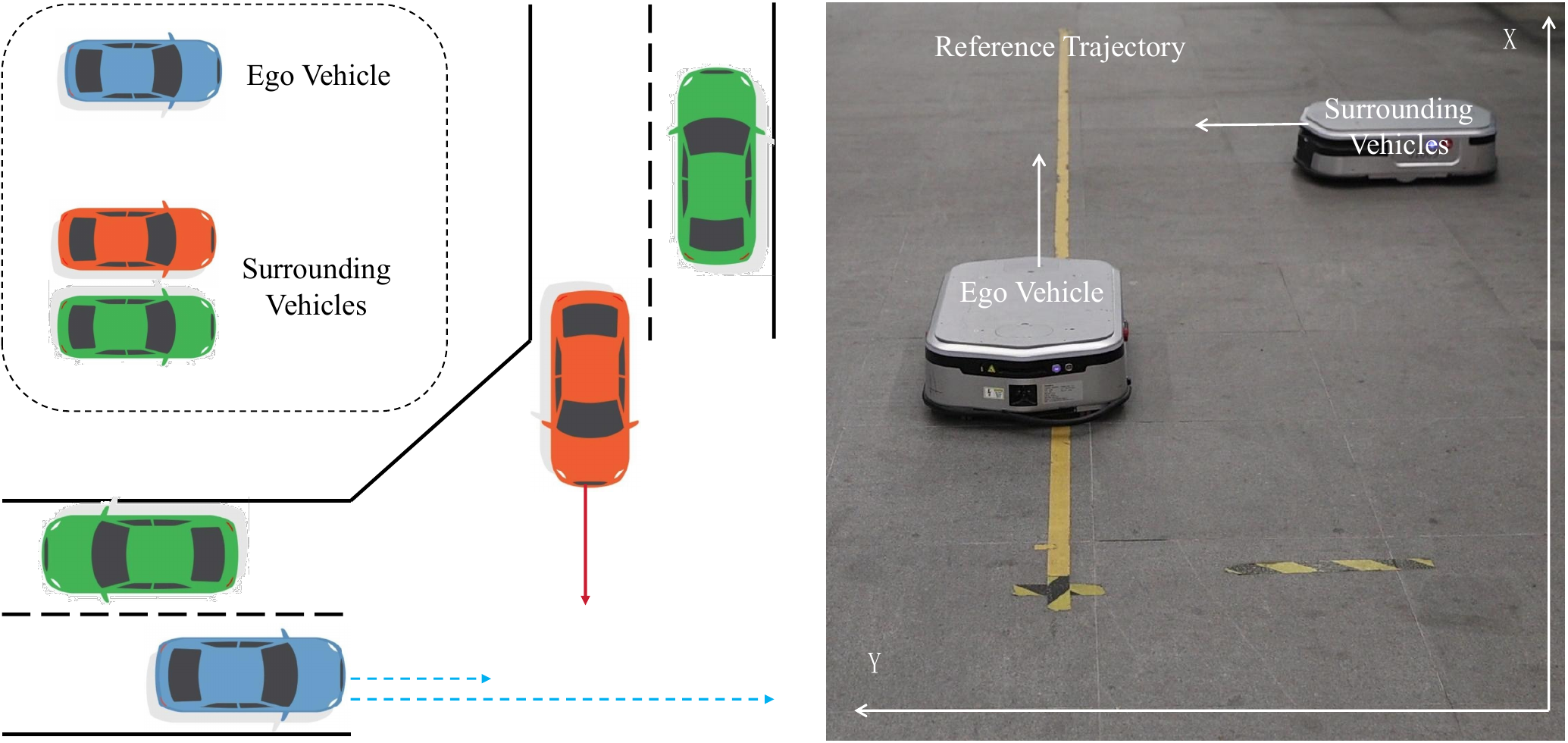}
        \caption{Experimental scenario 2}
        \label{Experimental Scene 2}
    \end{subfigure}

\caption{Schematic diagram of vehicle meeting environment}
\label{fig:vehicle_meeting_env}
\end{figure*}

This paper designs two vehicle tracking and collision avoidance tasks based on common driving scenarios: Scenario 1 is static obstacle avoidance and Scenario 2 is intersection obstacle avoidance, as described in Figure~\ref{fig:vehicle_meeting_env}. In the real-vehicle experiment, a small Automated Guided Vehicle (AGV) suitable for the application scenarios of unmanned warehouse logistics is selected.

\begin{figure*}[ht!]
    \centering
        \begin{subfigure}[b]{0.29\textwidth}
        \includegraphics[width=\textwidth]{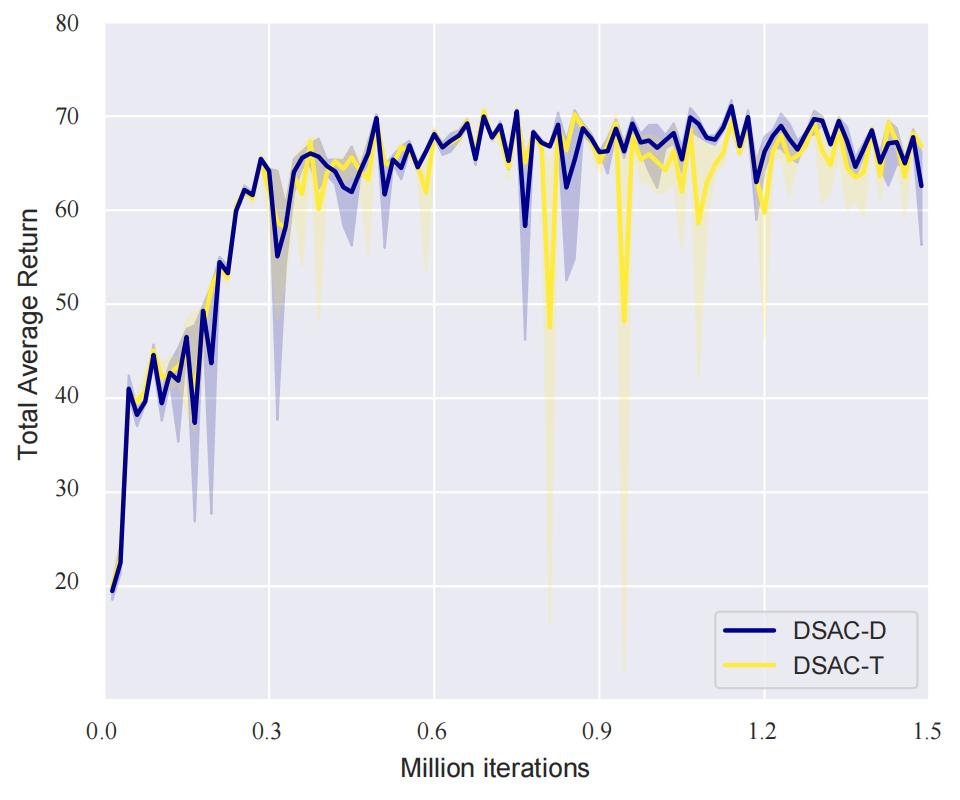}
        \caption{Comparison of algorithm performance}
        \label{trajectory_tar}
    \end{subfigure}
    \begin{subfigure}[b]{0.33\textwidth}
        \includegraphics[width=\textwidth]{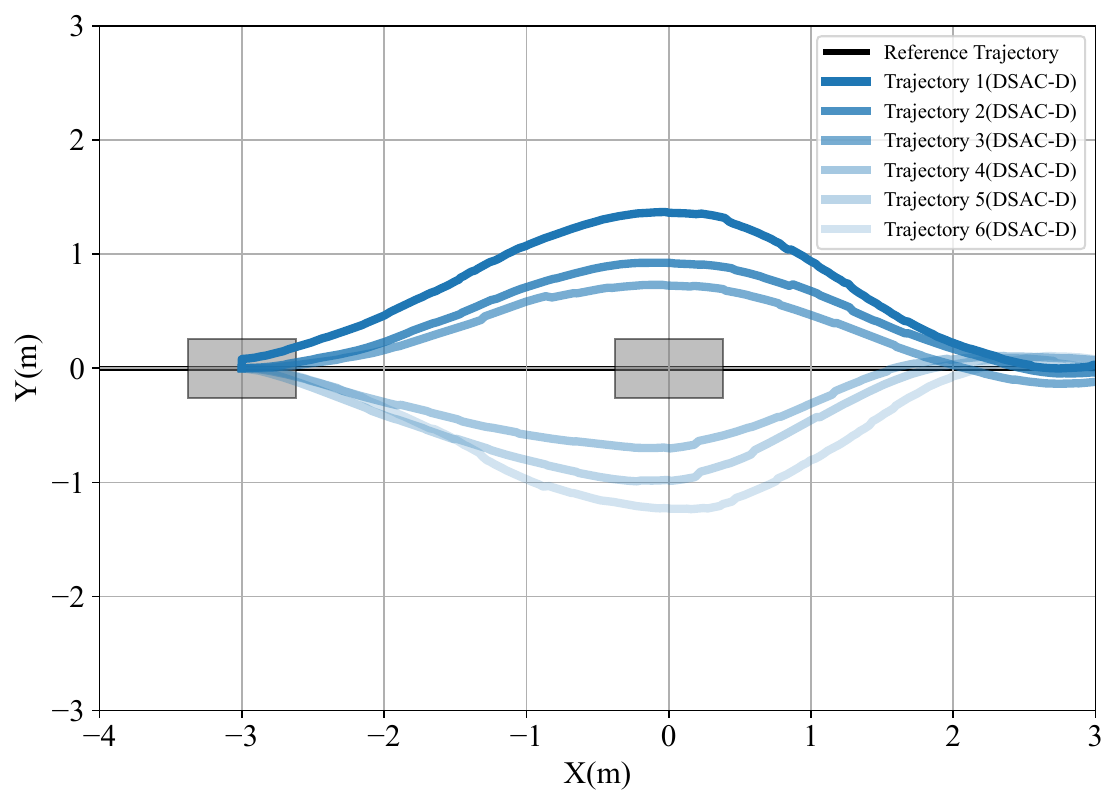}
        \caption{Multimodal trajectories of DSAC-D}
        \label{trajectory_dsacd}
    \end{subfigure}
    \begin{subfigure}[b]{0.33\textwidth}
        \includegraphics[width=\textwidth]{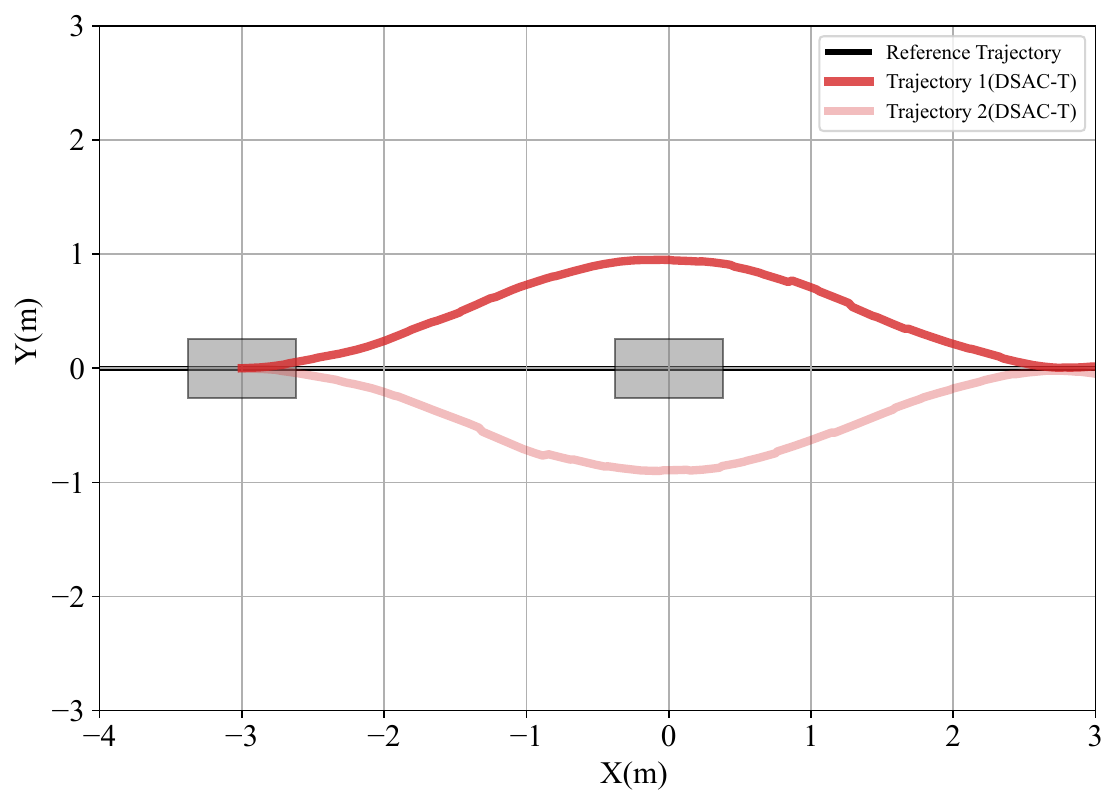}
        \caption{Multimodal trajectories of DSAC-T}
        \label{trajectory_dsact}
    \end{subfigure}
\caption{Training curve and multimodal trajectories in scenario 1}
\label{fig:vehicle_meeting_scene_1}
\end{figure*}

Using GOPS, we develop the diffusion reinforcement learning algorithm DSAC-D and deploy the DiffusionNet. We compare DSAC-D with DSAC-T in training and DiffusionNet with the DSAC-T's MLP policy network. Calculations show DSAC-D increases the average cumulative reward by 5.94\% compared to DSAC-T, as shown in Figure~\ref{trajectory_tar}. 

In the static obstacle avoidance scenario of Scenario 1, the DiffusionNet diffusion policy network trained by the DSAC-D algorithm can exhibit multiple trajectories of two types: left and right obstacle avoidance. Trajectories 1, 2, and 3 avoid from the left, keeping different distances from obstacles. Trajectories 4, 5, and 6 avoid from the right with the same effect, as in Figure~\ref{trajectory_dsacd}. In contrast, the MLP policy network trained by DSAC-T shows only single-modality trajectories, as in Figure~\ref{trajectory_dsact}. 

To verify the performance of DSAC-D algorithm under different multimodal distributional situations, an experiment is conducted in the relatively complex and dynamic scenario of the vehicle meeting task at a crossroads. Three different driving styles, namely aggressive, normal, and conservative, are set.   

\begin{figure*}[ht!]
    \centering
        \begin{subfigure}[b]{0.32\textwidth}
        \includegraphics[width=\textwidth]{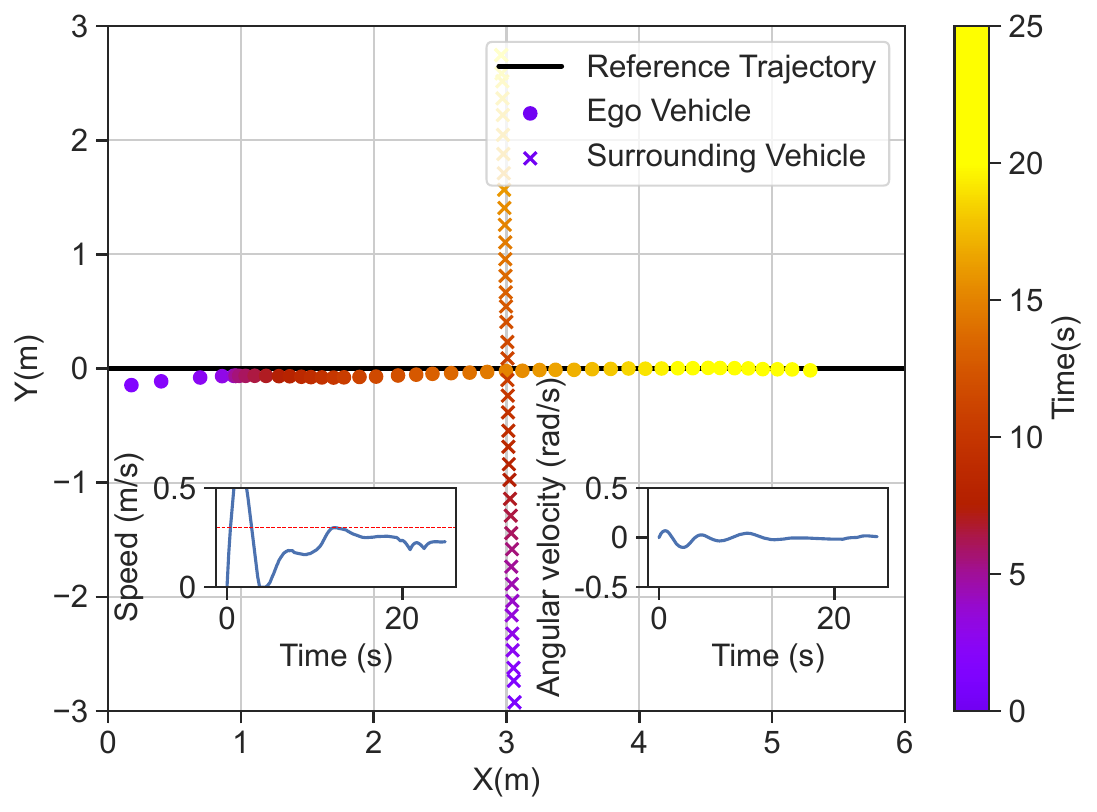}
        \caption{Trajectory of conservative driving style}
        \label{Trajectory of conservative driving style}
    \end{subfigure}
    \begin{subfigure}[b]{0.32\textwidth}
        \includegraphics[width=\textwidth]{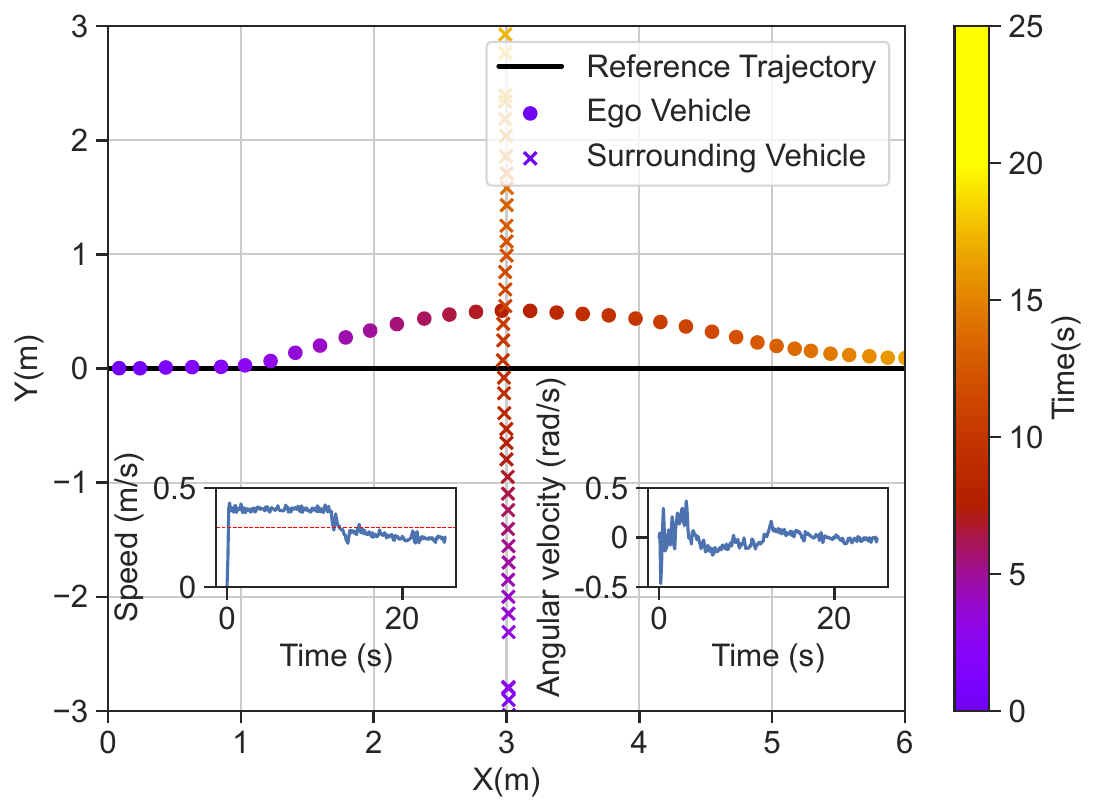}
        \caption{Trajectory of normal driving style}
        \label{Trajectory of normal driving style}
    \end{subfigure}
    \begin{subfigure}[b]{0.32\textwidth}
        \includegraphics[width=\textwidth]{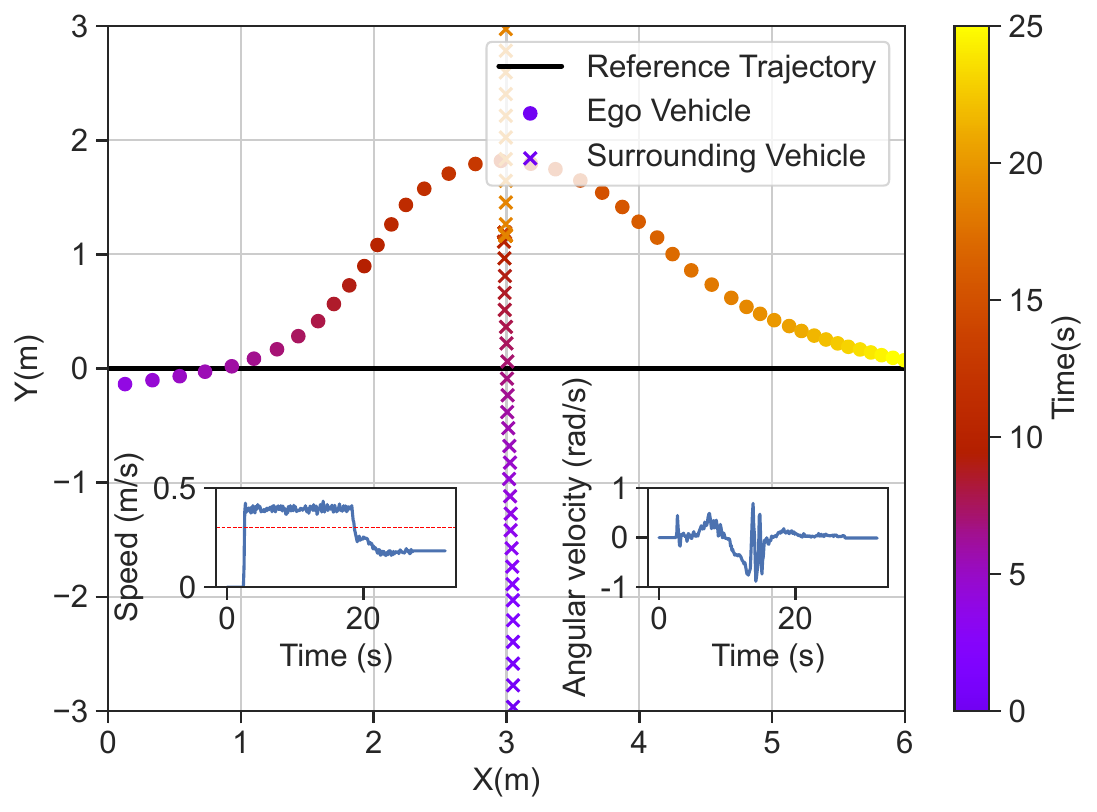}
        \caption{Trajectory of aggressive driving style}
        \label{Trajectory of aggressive driving style}
    \end{subfigure}

\caption{Multimodal trajectories with different driving styles in scenario 2}
\label{fig:vehicle_scene_2_tra}
\end{figure*}
\begin{figure*}[ht!]
    \centering
        \begin{subfigure}[b]{0.95\textwidth}
        \includegraphics[width=\textwidth]{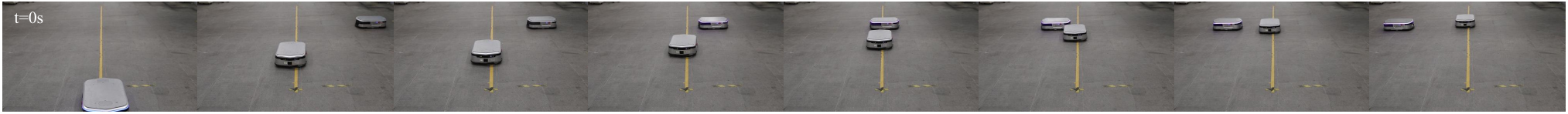}
        \caption{Experimental photos of conservative driving style}
        \label{Experimental photos of conservative driving style}
    \end{subfigure}
    \begin{subfigure}[b]{0.95\textwidth}
        \includegraphics[width=\textwidth]{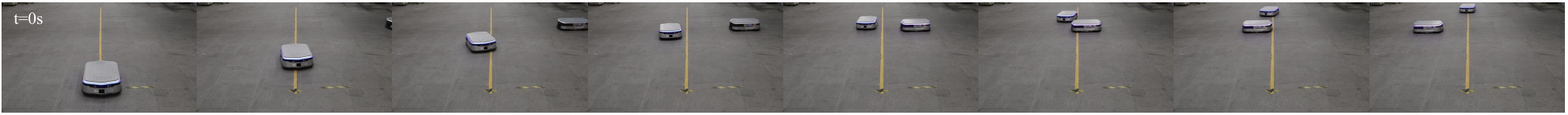}
        \caption{Experimental photos of normal driving style}
        \label{Experimental photos of normal driving style}
    \end{subfigure}
    \begin{subfigure}[b]{0.95\textwidth}
        \includegraphics[width=\textwidth]{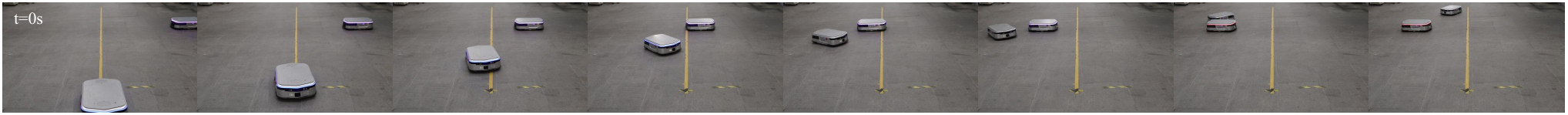}
        \caption{Experimental photos of aggressive driving style}
        \label{Experimental photos of aggressive driving style}
    \end{subfigure}

\caption{Experimental photos of different driving styles in scenario 2}
\label{Experimental photos of different driving styles in scenario 2}
\end{figure*}

Figure~\ref{fig:vehicle_scene_2_tra} shows the vehicle trajectories under three different driving styles. Figure~\ref{Experimental photos of different driving styles in scenario 2} presents the actual shooting pictures for these three styles. For the conservative, normal, and aggressive driving styles, the vehicle's steady-state trajectory tracking errors are 2.66 cm, 3.53 cm, and 4.98 cm respectively. Accounting for AGV sensor errors and experimental noise, DSAC-D's trajectory tracking errors in all styles at crossroads are under 5 cm. This demonstrates the high accuracy of the DSAC-D algorithm, allowing for precise completion of vehicle trajectory tracking and obstacle avoidance tasks.

\section{Discussion and Conclusions}
This paper introduces DSAC-D (Distributed Soft Actor Critic with Diffusion Policy), a distributional reinforcement learning algorithm designed to address the challenges of estimating bias in value functions and obtaining multimodal policy representations. By establishing a multimodal distribution policy iteration framework and leveraging a diffusion value network to accurately characterize reward sample distributions, DSAC-D effectively learns multimodal policies. Empirical results on both multi-objective and MuJoCo tasks demonstrate state-of-the-art performance across all nine control tasks. Notably, DSAC-D significantly reduces estimation bias and achieves a total average return improvement exceeding 10\% compared to existing mainstream algorithms. Furthermore, real vehicle testing confirms the algorithm's ability to accurately capture multimodal distributions reflective of varying driving styles, enabling the diffusion policy network to model multimodal trajectories effectively. These findings highlight the potential of DSAC-D for complex control tasks requiring nuanced understanding and representation of value distributions.

\end{document}